\documentclass[preprint,12pt]{elsarticle}

\usepackage{amsmath}
\usepackage{amssymb}
\usepackage{amsthm}
\usepackage{mathrsfs}
\usepackage{graphicx}
\usepackage{subfigure}
\usepackage[percent]{overpic}
\usepackage[hidelinks,breaklinks]{hyperref}
\usepackage[all]{xy}
\usepackage{algorithm}
\usepackage{algorithmic}
\usepackage{comment}
\usepackage{nth}
\usepackage{chngcntr}

\usepackage[margin=1in]{geometry}

\newcommand{\commentthis}[1]{{}}

\theoremstyle{plain}
\newtheorem{theorem}{Theorem}[section]

\theoremstyle{remark}
\newtheorem{definition}[theorem]{Definition}

\renewcommand{\P}{\mathbb{P}}
\newcommand{\tf}{\widetilde{f}}
\newcommand{\e}{\mathrm{e}}
\newcommand{\E}{\mathbb{E}}

\newcommand{\R}{\mathbb{R}}
\newcommand{\N}{\mathbb{N}}
\newcommand{\C}{\mathscr{C}}

\renewcommand{\d}{\mathrm{d}}
\renewcommand{\epsilon}{\varepsilon}

\newcommand{\dalpha}{D^{(\alpha)}_\epsilon}
\newcommand{\kalpha}{K^{(\alpha)}_\epsilon}
\newcommand{\malpha}{M^{(\alpha)}_\epsilon}
\newcommand{\palpha}{P^{(\alpha)}_\epsilon}
\newcommand{\bfxi}{\xi}
\newcommand{\tildeXi}{\widetilde{\bfxi}}
\newcommand{\tildexi}{\widetilde{\xi}}

\DeclareMathOperator*{\argmin}{arg\,min}
\DeclareMathOperator*{\dist}{dist}

\newcommand{\secref}[1]{Section~\ref{#1}}

\newcommand{\todo}[1]{\footnote{\footnotesize \textbf{To do:} {#1}}}

\usepackage[colornames]{xcolor}

\DeclareMathOperator{\id}{id}

\numberwithin{equation}{section}

\definecolor{rubyglass}{HTML}{B3444B}

\journal{Applied and Computational Harmonic Analysis}

\begin{document}



\begin{frontmatter}



\title{On the Correspondence\\between Gaussian Processes and Geometric Harmonics}


\author[labeltum]{Felix Dietrich}
\ead{felix.dietrich@tum.de}

\affiliation[labeltum]{organization={Department of Informatics, Technical University of Munich}, addressline={Boltzmannstr. 3}, city={Garching at Munich}, postcode={85748}, state={Bavaria}, country={Germany}}

\author[labeljhu]{Juan M. Bello-Rivas}
\author[labeljhu]{Ioannis G. Kevrekidis}

\affiliation[labeljhu]{organization={Department of Chemical and Biomolecular Engineering, Whiting School of Engineering, Johns Hopkins University}, addressline={3400 North Charles Street}, city={Baltimore}, postcode={21218}, state={MD}, country={USA}}

\begin{abstract}
    We discuss the correspondence between Gaussian process regression and Geometric Harmonics, two similar kernel-based methods that are typically used in different contexts.
    Research communities surrounding the two concepts often pursue different goals.
    Results from both camps can be successfully combined,
    providing alternative interpretations of uncertainty in terms of error estimation, or leading towards accelerated Bayesian Optimization due to dimensionality reduction.
\end{abstract}



\begin{keyword}
Gaussian Processes \sep Geometric Harmonics \sep Uncertainty \sep Dimension reduction \sep Bayesian optimization

\MSC 42-XX \sep 42-08 \sep 60G15

\end{keyword}

\end{frontmatter}

\section{Introduction}
\label{sec:introduction}

Gaussian Processes~\cite{Adler2007a} (GPs) are simultaneously an important class of stochastic processes, with Brownian motion being the most prominent example, and a practical engineering tool~\cite{Rasmussen2006} used e.g. for regression, having first appeared as the Kriging method in the geosciences.
The main use of Gaussian Process Regression is to provide a surrogate model to make predictions based on empirical data.

Geometric Harmonics~\cite{lafon_diffusion_2004,coifman_geometric_2006} (GHs) constitute an out-of-sample extension method that originates in the computational harmonic analysis literature, and are related to non-linear dimensionality reduction, Fourier series, and the Nystr\"om extension~\cite{williams_using_2000,bengio_out--sample_2003}.
Given a real-valued function defined on a discrete set of data points, GHs (like GPs) allow us to evaluate the function on a point not contained in the original data set.

There is a large body of literature on spline interpolation in statistics, dealing with the connection of Reproducing Kernel Hilbert Spaces (RHKS) and Bayesian statistics~\cite{wahba-1990}. In many cases, the convergence rates of spline interpolation for an unknown function can be related to the rate of decay of the eigenvalues of the reproducing kernel, and of the Fourier-Bessel coefficients of the function being estimated with respect to the eigenfunctions.
This is directly related to Geometric Harmonics, and their interpretation as generalized Fourier basis expansion of the function.

Despite the immediate similarity between the goals of Gaussian Process Regression and Geometric Harmonics, there appears to be fundamental distinctions between the use of two methods.
Geometric Harmonics are often used {\em after identifying a lower dimensional manifold} on which the data points lie and, for this reason, can be a less expensive computational technique.
They are also used for applications such as de-noising, which is not the primary use of Gaussian Process Regression.

\sloppy The similarity of the two approaches---particularly, the method of Kriging and Geometric Harmonics---has been noted, both in the original paper by Coifman et al.~\cite{coifman_geometric_2006} and by others~\cite{barkan-2016}. These authors mostly focused on the noise-free interpretation of GH.
There has been interest in studying Gaussian Processes in latent spaces~\cite{lawrence_gaussian_2004,lawrence_probabilistic_2005,titsias_bayesian_2010,thimmisetty-2017,Soize-2021,Soize-2021a}, even in latent spaces constructed---as the initial step---through manifold learning.
To the best of our knowledge, these efforts do not take advantage of the similarity of the kernels used for manifold learning, and for subsequent regression on this manifold; here we advocate the iterative, on-line use of the same kernel for non-linear manifold learning {\em and} regression.

This link between Gaussian Process Regression and Geometric Harmonics
allows us to exploit specific strengths of each method: uncertainty quantification and established software tool chains in the case of Gaussian Process Regression as well as dimensionality reduction and error estimates in the case of Geometric Harmonics.
We posit the usefulness of merging both approaches to tackle iterative computational tasks  in problems for which an underlying low-dimensional manifold is relevant.

\begin{figure}[ht!]
  \centering 
  \subfigure[Schematic and computations of a few reduced optimization steps. ]{
    \includegraphics[width=0.5855\columnwidth]{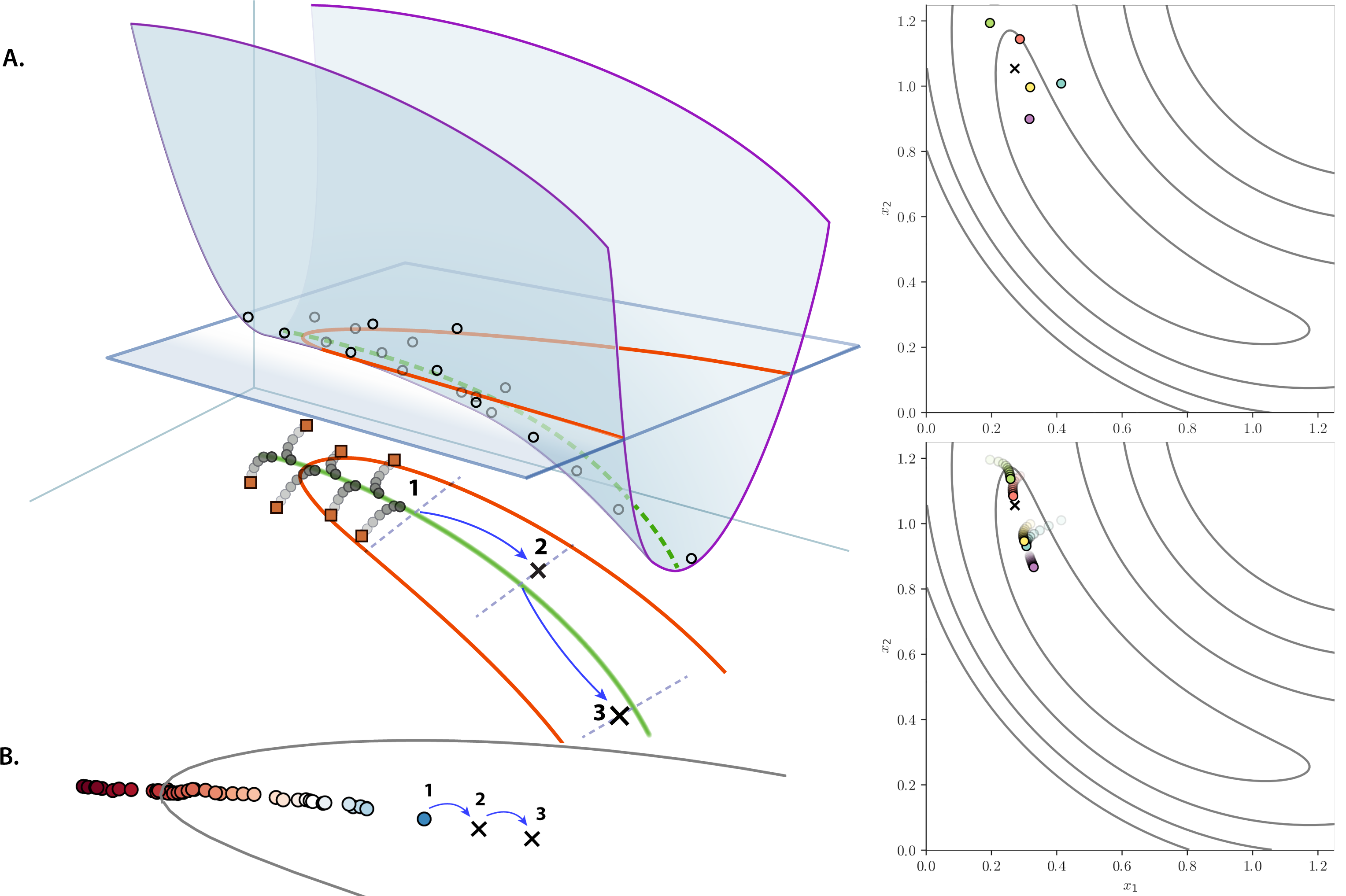}
  } \\
  \subfigure[Step 1 (dim. reduction)]{
    \begin{overpic}[width=0.23\columnwidth,trim=1.45cm 1.75cm 1.5cm 1.5cm,clip]{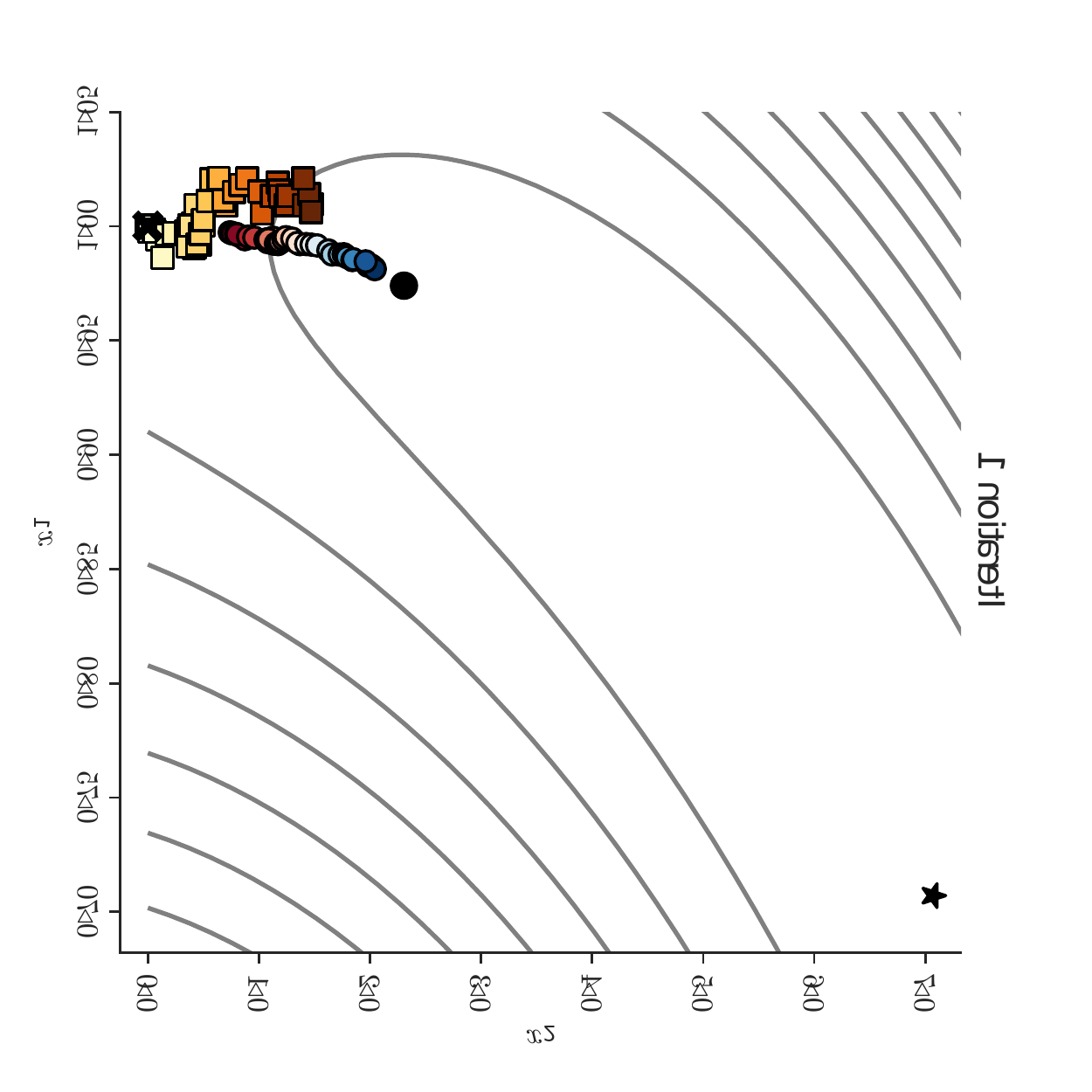}
      \put(-3, 91){\footnotesize{\nth{0}}}
      \put(30, 82.5){\footnotesize{\nth{1}}}
      \put(68, 10){\footnotesize{Optimum}}
    \end{overpic}
  }
  \subfigure[Step 2 (dim. reduction)]{
    \begin{overpic}[width=0.23\columnwidth,trim=1.45cm 7cm 6cm 1cm,clip]{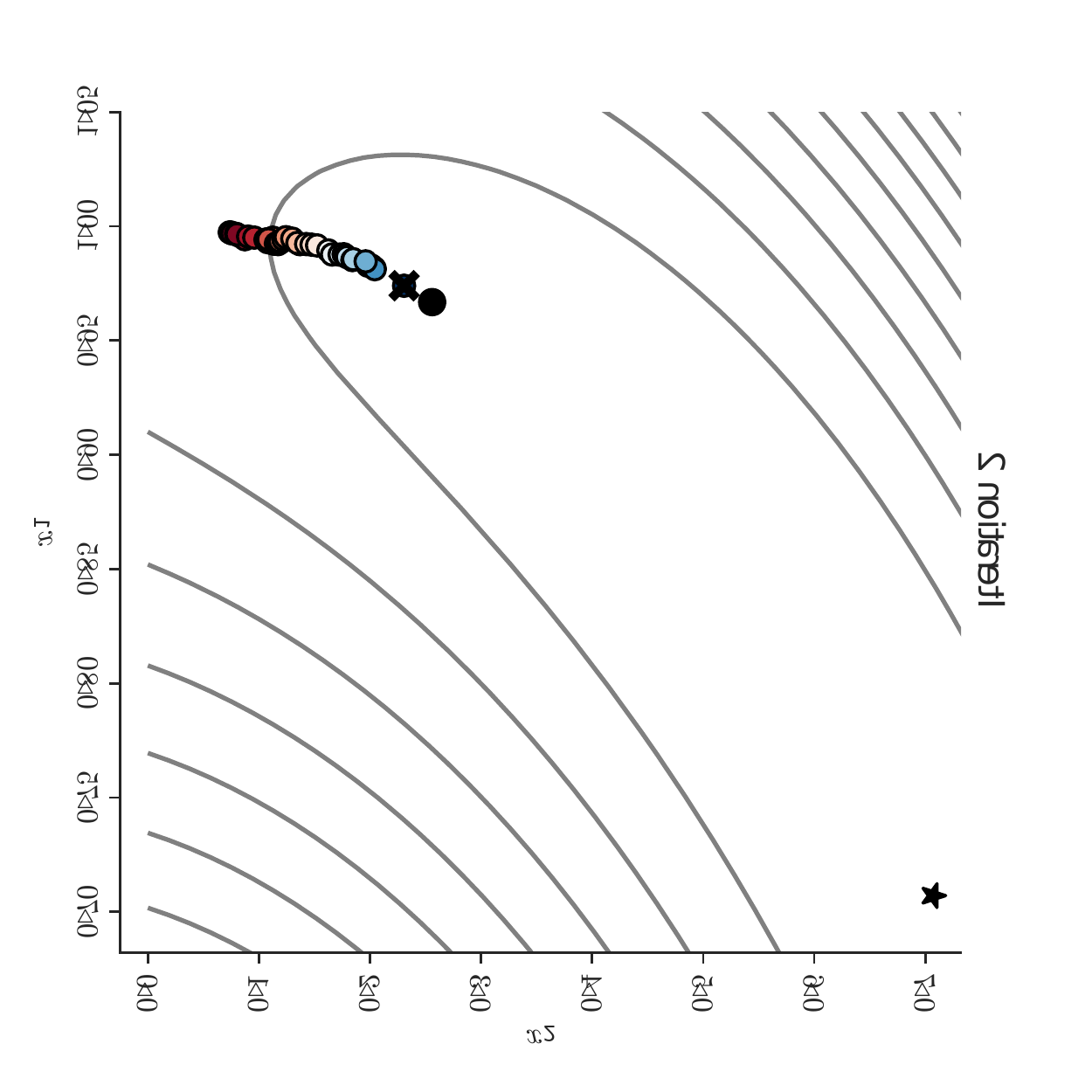}
      \put(60, 55){\nth{1}}
      \put(67.5, 47.5){\nth{2}}
    \end{overpic}
  }
  \subfigure[Step 3 (dim. reduction)]{
    \begin{overpic}[width=0.23\columnwidth,trim=1.45cm 7cm 6cm 1cm,clip]{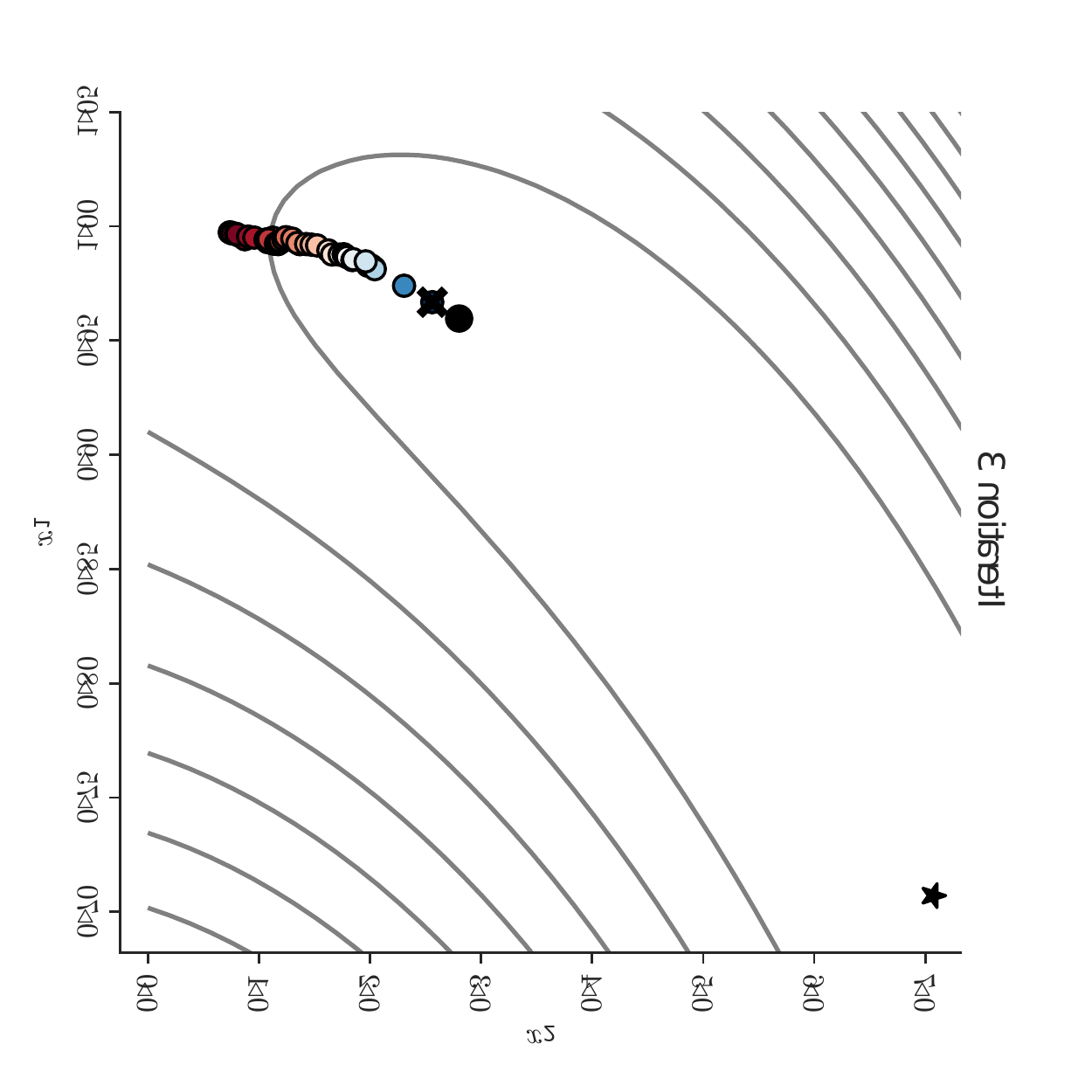}
      \put(67.5, 47.5){\nth{2}}
      \put(77.5, 40){\nth{3}}
    \end{overpic}
  }\\
  \subfigure[Step 1 (plain)]{
    \begin{overpic}[width=0.23\columnwidth,trim=1.45cm 1.75cm 1.5cm 1.5cm,clip]{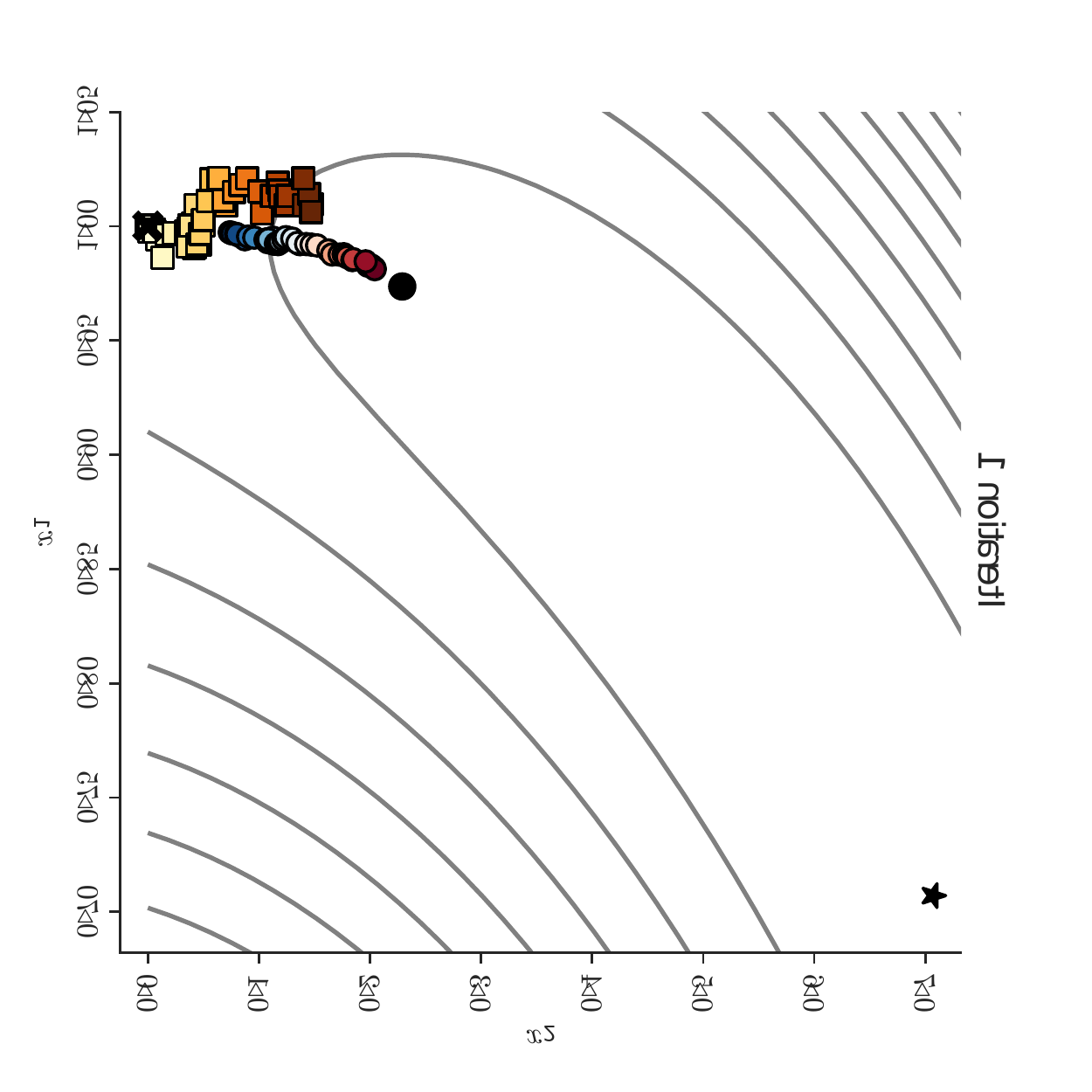}
      \put(-3, 91){\footnotesize{\nth{0}}}
      \put(30, 82.5){\footnotesize{\nth{1}}}
      \put(68, 10){\footnotesize{Optimum}}
    \end{overpic}
  }
  \subfigure[Step 2 (plain)]{
    \begin{overpic}[width=0.23\columnwidth,trim=1.45cm 7cm 6cm 1cm,clip]{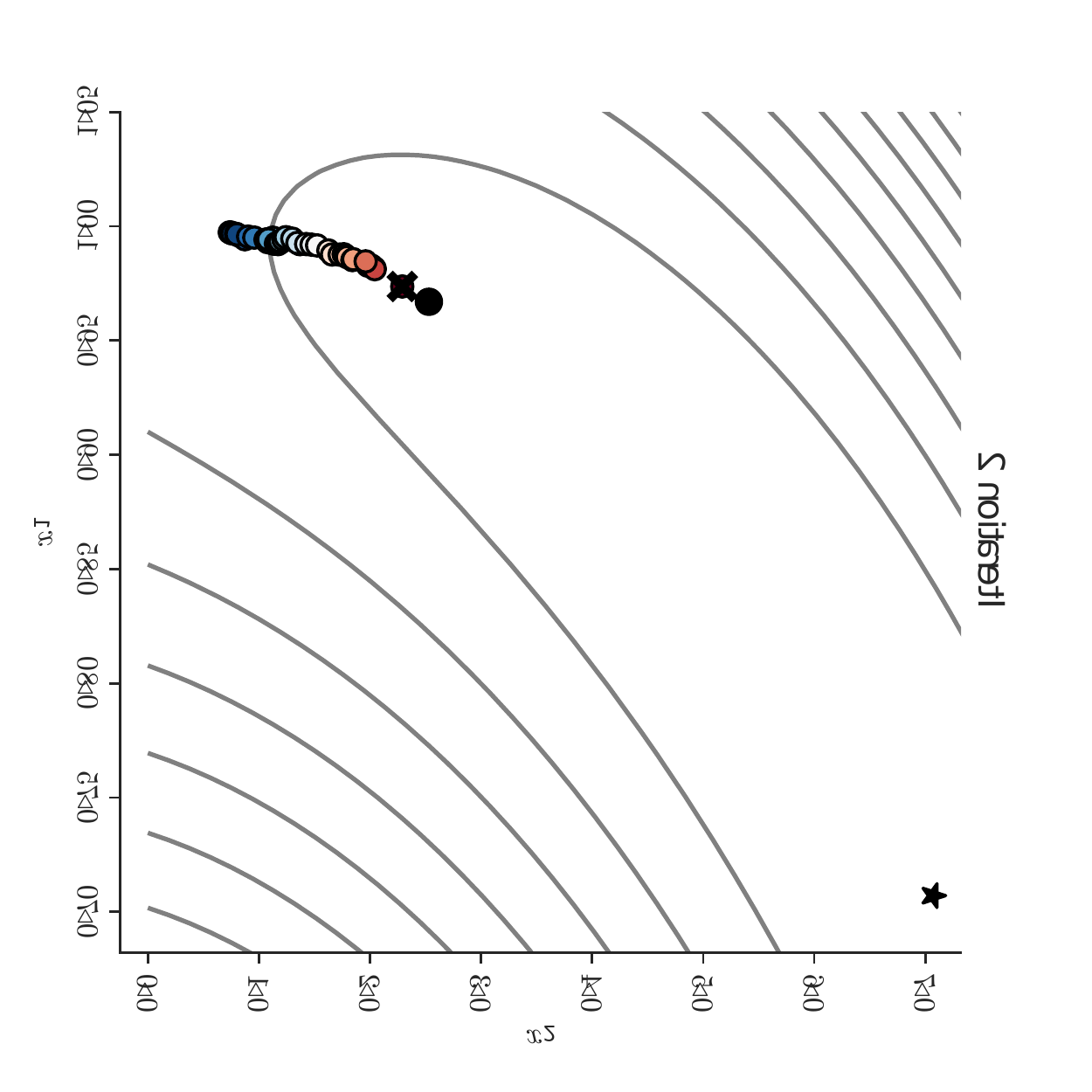}
      \put(60, 55){\nth{1}}
      \put(67.5, 47.5){\nth{2}}
    \end{overpic}
  }
  \subfigure[Step 3 (plain)]{
    \begin{overpic}[width=0.23\columnwidth,trim=1.45cm 7cm 6cm 1cm,clip]{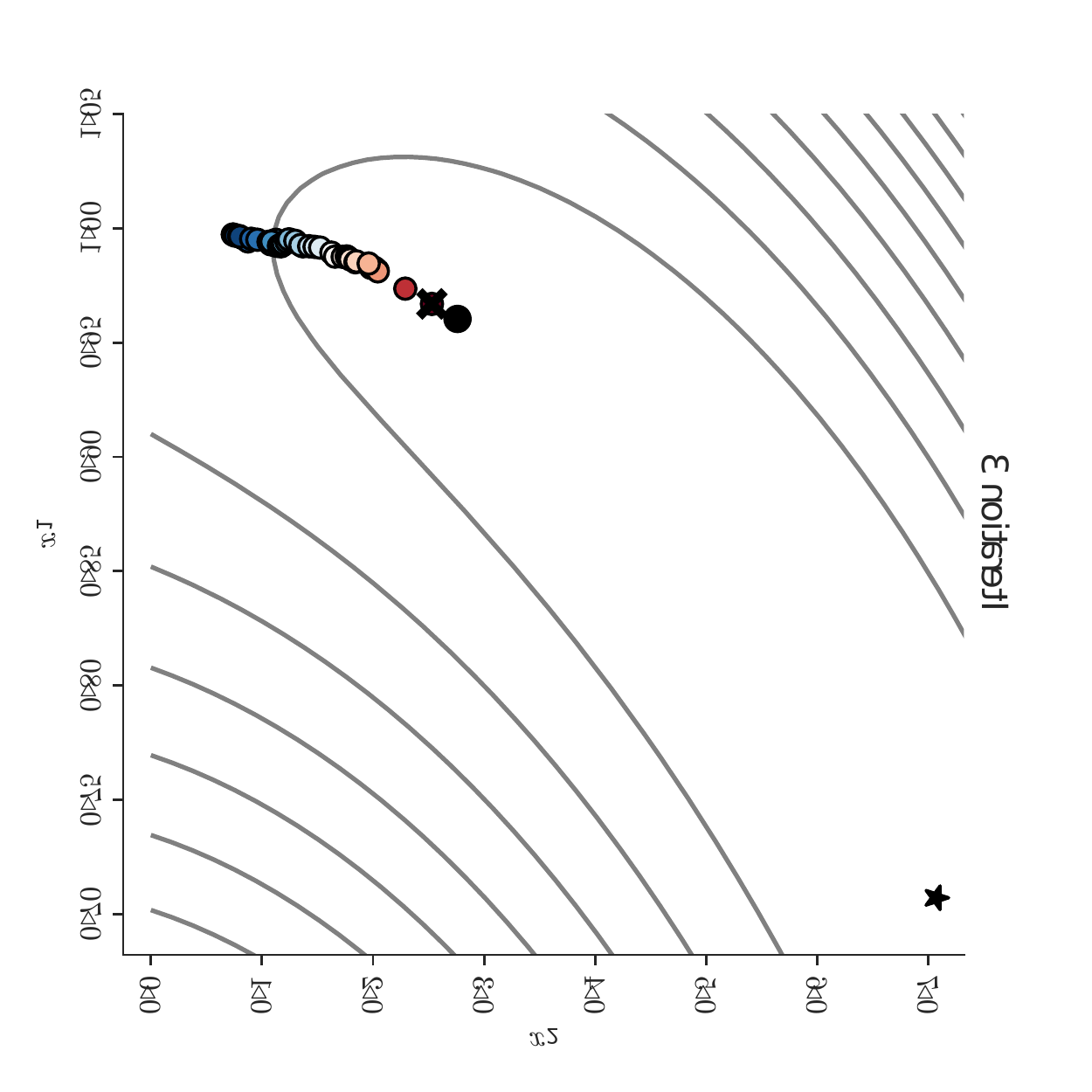}
      \put(67.5, 47.5){\nth{2}}
      \put(77.5, 40){\nth{3}}
    \end{overpic}
  }
  \caption{Dimensionally reduced optimization. (A, top left) Schematic rendering of a multiscale optimization problem, with a canyon-shaped objective function. Starting from an initial sample, a few steps of gradient descent suffice to relax each of the sampled points to a one-dimensional submanifold, the bottom of the canyon. Projections of actual computations (sampling, above, and relaxation, below) are shown at the top right. (B) a couple of subsequent ``low-dimensional" optimization steps. Iterations of Bayesian optimization using dimensionality reduction (Middle, b-d) and plain Bayesian optimization in the original space (Bottom, e-g). The initial condition at each iteration is marked with a cross ($\times$), The points sampled by Metropolis-Hastings are depicted by squares ($\square$), the points after relaxation are represented by dots ($\circ$), and the lifted optimum at the current iteration is shown as a black dot ($\bullet$). The global optimum is marked with a star ($\star$).}
  \label{fig:canyon-artistic}
\end{figure}

A textbook example for the relevance of a low-dimensional manifold in modeling is the case of singularly perturbed systems of ordinary differential equations: after a short transient, the dynamics converge to such a reduced, ``slow manifold''. There is a direct analogy with gradient descent {\em dynamics} in a multi-scale  optimization problem, where the objective function is very steep in most directions and slowly changing in the few remaining ones. Here, it is the {\em dynamics of the optimization algorithm} that ultimately evolve on a reduced slow manifold (e.g. \cite{Pozharskiy-2020}).
Figure~\ref{fig:canyon-artistic} provides a schematic of such a multiscale optimization, along with a few steps of its realization in a simple illustrative example.
A few initial steps of gradient descent suffice to relax each of the sampled points (see Figure~\ref{fig:canyon-artistic}) to a conceptual one-dimensional ``bottom of the canyon''. In such a setting, it could be advantageous to combine dimensionality reduction with regression in order to build progressive surrogate models in a dimensionally reduced space, thus aspiring to decrease the computational expense of the optimization procedure, e.g. the exploitation/exploration steps in a Bayesian optimization (BO) scheme. Importantly, this will require a systematic {\em lifting} procedure: reliably going back and forth between the reduced and the ``full'' spaces; the same tools will be useful in many other computational tasks beyond optimization (reduced simulation, reduced fixed point computations, reduced controller design, see \cite{theodoropoulos-2000,kevrekidis-2003,kevrekidis-2009,Arbabi-2020}). 
The model and computations leading to illustrative panels in Figure~\ref{fig:canyon-artistic} are described in the Appendix.

The remainder of the paper is organized as follows:
In \secref{sec:mathematical setting} we summarize the mathematical background of the Karhunen-Lo\`{e}ve decomposition as a basis for both GP and GH.
In \secref{sec:uncertainty}, we show that the formal correspondence allows us to merge results from GHs and GPs, in particular their respective error and uncertainty estimates. We discuss some inductive biases of GPs through the lens of GHs.
We tackle numerical algorithms for both approaches in \secref{sec:numerical algorithms}, again illustrating direct correspondence.
In \secref{sec:dimension reduction}, we outline how function approximation can be done on latent spaces which are immediately accessible from previous GP and GH computations.
We conclude in \secref{sec:discussion} with a summary and a discussion of possible research directions.

In \ref{sec:towards-reduced-bayesian-optimization}, we revisit our simple example, showing how the correspondence between GPs and GHs in the case of a ``slow'' one-dimensional manifold embedded in two dimensions.

\section{The mathematical setting}
\label{sec:mathematical setting}

We now recall the formulation of Gaussian processes and geometric harmonics and relate the two via the Karhunen-Lo\`{e}ve expansion.
We start with the statistical setting and define the term Gaussian (random) field---a generalized Gaussian process---in \secref{sec:gaussian fields}. We state that these fields can be expanded as a series of eigenfunctions of the covariance kernel in \secref{sec:kl expansion}.

With these prerequisites, we recall how low-dimensional descriptions can be obtained directly from these eigenfunctions in \secref{sec:dmap and gpr}, bridging the gap from statistics to differential geometry and functional analysis. In particular, a Gaussian process is related to the Laplace-Beltrami operator if the kernel employed in it is normalized in a specific way (as outlined by the Diffusion Maps algorithm).
This normalization is discussed in \secref{sec:kernels}.

\subsection{Gaussian fields}
\label{sec:gaussian fields}

\begin{definition}
  The $\R^d$-valued random variable $\xi = (\xi_1, \dotsc, \xi_d)$ is said to be a multivariate Gaussian with mean $\mu$ and covariance $C$, denoted by $\xi \sim N(\mu, C)$, if, for every $\alpha \in \R^d$, the random variable $\langle \alpha, \xi \rangle = \sum_{i = 1}^d \alpha_i \xi_i$ is Gaussian.
  The mean vector $\mu \in \R^d$ and covariance matrix $C \in \R^{d \times d}$ of $\xi$ are respectively given by the components
  \begin{equation*}
    \mu_i = \E[\xi_i]
    \qquad \text{and} \qquad
    C_{ij} = \E[ (\xi_i - \mu_i) (\xi_j - \mu_j) ],
  \end{equation*}
  for $i, j \in \{ 1, \dotsc, d \}$.
  The probability density of $\xi$ is given by
  \begin{equation*}
    p(z)
    =
    \frac{1}{\sqrt{(2 \pi)^d \det C}}
    \exp\left\{ -\tfrac{1}{2} (z - \mu)^\top C^{-1} (z - \mu) \right\}.
  \end{equation*}
\end{definition}

\begin{definition}
  Let $(\Omega, \Sigma, \mathbb{P})$ be a probability space and let $T \subseteq \R^k$ be a compact set.
  The collection of random variables $\{ f(x) \in \R^d \, | \, x \in T \}$ is called a (vector-valued) random field.
\end{definition}

\begin{definition}
  A Gaussian random field is a random field $\{ f(x) \, | \, x \in T \}$ determined by the mean and covariance functions
  \begin{equation*}
    \mu(x) = \E[f(x)]
    \qquad \text{and} \qquad
    C(x, y) = \E[ (f(x) - \mu(x)) (f(x) - \mu(y)) ].
  \end{equation*}
  An $\R^d$-valued multivariate Gaussian field $\{ f(x) = (f_1(x), \dotsc, f_d(x)) \, | \, x \in T \}$ is analogously characterized by
  \begin{equation*}
    \mu_i(x) = \E[f_i(x)]
    \qquad \text{and} \qquad
    C_{ij}(x, y) = \E[ (f_i(x) - \mu_i(x)) (f_j(x) - \mu_j(y)) ],
  \end{equation*}
  for $i, j \in \{ 1, \dotsc, d \}$.
\end{definition}

\subsection{The Karhunen-Lo\`eve expansion}\label{sec:kl expansion}

Every centered Gaussian process $f$ with a continuous covariance function $C$ admits an expansion of the form
\begin{equation}
  \label{eq:onb-expansion}
  f(x, \omega)
  =
  \sum_{n = 1}^\infty
  \kappa_n(x)
  \,
  \xi_n(\omega),
\end{equation}
where $\{ \xi_n(\omega) \sim N(0, 1) \, | \, n \in \N \}$
and the functions $\{ \kappa_n \colon M \to \R \, | \, n \in \N \}$ are determined by the covariance $C$ in a manner that we will see next.
When $M \subset \R^m$ is compact, the formula~\eqref{eq:onb-expansion} is known as the Karhunen-Lo\`eve decomposition~\cite{Adler2007a,Adler2007b}.

Let us define the operator $\C$ as
\begin{equation*}
  (\C \phi)(x)
  =
  \int_M C(x, y) \, \phi(y) \, \d y
\end{equation*}
The eigenpairs $\{ (\lambda_n, \phi_n) \, | \, n \in \N \}$ of $\C$ are such that
\begin{enumerate}
\item $(\C \phi_n)(x) = \lambda_n \phi_n(x)$.
\item $\lambda_1 \ge \lambda_2 \ge \dotsb > 0$.
\item $\int_M \phi_n(x) \phi_m(x) \, \d x = \delta_{n, m}$ for $n, m \in \N$.
\end{enumerate}
By Mercer's theorem~\cite{mercer1909}, we know that
\begin{equation*}
  C(x, y)
  =
  \sum_{n \in \N}
  \lambda_n \, \phi_n(x) \phi_n(y),
\end{equation*}
where the series converges uniformly and absolutely on $M \times M$.
The Karhunen-Lo\`eve expansion of $f$ is given by
\begin{equation}
  \label{eq:kl-expansion}
  f(x, \omega)
  =
  \sum_{n \in \N}
  \sqrt{\lambda_n} \,
  \phi_n(x) \,
  \xi_n(\omega),
\end{equation}
where $\xi_n \sim N(0, 1)$.
After writing $\kappa_n = \sqrt{\lambda_n} \phi_n$, we see that~\eqref{eq:kl-expansion} becomes~\eqref{eq:onb-expansion}.

\subsection{Diffusion maps and Gaussian process regression}\label{sec:dmap and gpr}

In this section we establish a link between diffusion maps~\cite{Coifman2006} and Gaussian process regression~\cite{Rasmussen2006} using the conditional Karhunen-Lo\`eve expansion~\cite{Adler2007a,Ossiander2014,Tipireddy2019}.
This approach enables the study of the following topics from a different viewpoint than the usual one in which their approached:
\begin{enumerate}
\item Quantifying uncertainty in the context of diffusion maps.
\item Studying deep Gaussian processes from a geometric standpoint.
\item Transferring the notion of optimality (maximum likelihood) from the Gaussian process side to the diffusion maps side.
\item Obtaining diffusion maps from Gaussian process regression.
\end{enumerate}

Let $M$ be a compact Riemannian manifold of dimension $m$.
We denote the geodesic distance between two points $x, y \in M$ by $d(x, y)$.
We regard $M$ as a probability space with an absolutely continuous probability measure $\P$.

\subsubsection{Kernels}
\label{sec:kernels}

Let $\epsilon > 0$.
Consider the kernel $K_\epsilon \colon M \times M \to \R_{\ge 0}$ given by
\begin{equation}
  (x, y)
  \mapsto
  K_\epsilon(x, y)
  =
  \e^{ -\frac{d(x, y)^2}{2 \epsilon} },
\end{equation}
where $d$ is the geodesic distance between points $x$ and $y$ in $M$.
Let $\alpha \in [0, 1] \subset \R$ and let
\begin{equation}
  Q_\epsilon(x) = \int K_\epsilon(x, y) \, \d \P(y).
\end{equation}
Now we construct the renormalized kernel
\begin{equation*}
  \kalpha(x, y)
  =
  \frac{K_\epsilon(x, y)}{Q_\epsilon(x)^\alpha \,  Q_\epsilon(y)^\alpha}.
\end{equation*}
Applying the weighted graph Laplacian normalization to $\kalpha$ above,
\begin{equation*}
  \dalpha(x)
  =
  \int \kalpha(x, y) \, \d \P(y)
\end{equation*}
we define the Markov kernel
\begin{equation*}
  \palpha(x, y)
  =
  \frac{\kalpha(x, y)}{\dalpha(x)}.
\end{equation*}
Note that the above kernel is not symmetric, however, it is conjugate to the symmetric kernel
\begin{equation*}
  \malpha(x, y)
  =
  \frac{\kalpha(x, y)}{\sqrt{\dalpha(x) \, \dalpha(y)}}.
\end{equation*}

The kernel $P^{(\alpha)}_\epsilon$ is positive semi-definite, as it is the transition kernel of a Markov operator~\cite{nummelin1984}.
This implies that $M^{(\alpha)}_\epsilon$ is symmetric and positive semi-definite.
Therefore, by Mercer's theorem, the kernel $\malpha$ admits the spectral decomposition
\begin{equation*}
  \malpha(x, y)
  =
  \sum_{n \in \N} \lambda_n \, \phi_n(x) \, \phi_n(y),
\end{equation*}
where $(\lambda_n, \phi_n)$ is an eigenpair of the linear operator defined by
\begin{equation*}
  \mathscr{M}^{(\alpha)}_\epsilon h
  =
  \int
  \malpha(\cdot, y) \, h(y) \, \d \P(y).
\end{equation*}

Note that if $(\lambda, \phi)$ is an eigenpair of $\mathscr{M}^{(\alpha)}_\epsilon$, then
$\lambda$ is also an eigenvalue of the operator $\mathscr{P}^{(\alpha)}_\epsilon$ defined by
\begin{equation*}
  \mathscr{P}^{(\alpha)}_\epsilon h
  =
  \int \palpha(\cdot, y) \, h(y) \, \d \P(y)
\end{equation*}
with eigenfunction
\begin{equation}
  \label{eq:relation-eigenfunctions}
  \psi
  =
  \frac{\phi}{\sqrt{\dalpha}}.
\end{equation}

\subsubsection{Conditional Karhunen-Lo\`eve expansion}\label{sec:conditional kl expansion}

We now adopt the extrinsic viewpoint and regard the $m$-dimensional Riemannian manifold $M$ as being embedded in the Euclidean space $\R^n$ where $m \le n$.
In particular, the eigenfunctions $\phi_n$ are understood to be extensions to $\R^n$ of the eigenfunctions of the Laplace-Beltrami operator on $M$ (see~\cite[Section 4.2]{coifman_geometric_2006}).

Let $f \colon M \subset \R^n \to \R$ be an integrable function
By the Karhunen-Lo\`eve decomposition, we have
\begin{equation}
  \label{eq:f-series}
  f(x, \omega)
  =
  \sum_{n = 1}^\infty \sqrt{\lambda_n} \, \phi_n(x) \, \xi_n(\omega),
\end{equation}
where $\xi_n \sim N(0, 1)$ and $\{ (\lambda_n, \phi_n) \, | \, n \in \N \}$ are eigenpairs of the covariance matrix $M$.

In what follows, we abuse notation and proceed to define vectors and matrices with countably many components so that the expressions adopt a more concise form.
However, these objects will be suitably truncated later.

Consider the set of points $\{ x_1, \dotsc, x_N \in M \} \subset \R^n$.
Let $\Lambda$ be the diagonal matrix with diagonal entries $\lambda_i$ for $i \in \N$, let $\xi = (\xi_1(\omega), \xi_2(\omega), \dotsc)$ and let $\Phi$ be the matrix whose $(i, j)$-th entry is equal to $\phi_i(x_j)$ for $i \in \N$ and $j = 1, \dotsc, N$.
We can rewrite~\eqref{eq:f-series} in the more compact notation
\begin{equation}
  \label{eq:f-matrices}
  f
  =
  \Phi^\top \, \Lambda^{1/2} \, \xi.
\end{equation}

Now consider the Gaussian random vector $Y = (Y_1, \dotsc, Y_N)$, where $Y_j = f(x_j, \omega) + \eta_j$ for $j = 1, \dotsc, N$, where $\eta_j$ are independent Gaussians with mean zero and variance $\sigma^2$.
Conditioning $Y$ on the event $\{ Y = y \}$, where $y = (y_1, \dotsc, y_N) \in \R^N$, leads us to a Gaussian process~\cite{Ossiander2014,Tipireddy2019} that can be expressed as
\begin{equation}
  \label{eq:f-tilde}
  \tf(x, \omega)
  =
  \Phi^\top \Lambda^{1/2} \, \tildeXi(\omega).
\end{equation}
where $\tildeXi$ is a Gaussian random vector resulting from conditioning $\xi$.
More concretely, conditioning the Gaussian random vector $(\xi, Y)$ on $\{ Y = y \}$ and using the identities
\begin{equation*}
  \E[ \xi ] = \E[ Y ] = 0,
  \quad
  \E[ \xi \, Y^\top ] = \Lambda^{1/2} \Phi,
  \quad \text{and} \quad
  \E[ Y \, Y^\top ] = M + \sigma^2 I,
\end{equation*}
we conclude that $\tildeXi = \xi |_{Y = y}$ has mean $\mu_{\tildeXi}$ and covariance matrix $C_{\tildeXi}$ respectively given by
\begin{equation}
  \label{eq:conditional}
  \mu_{\tildeXi}
  =
  \Lambda^{1/2} \Phi (M + \sigma^2 I)^{-1} y
  \quad \text{and} \quad
  C_{\tildeXi}
  =
  I
  -
  \Lambda^{1/2} \Phi ( M + \sigma^2 I)^{-1} \Phi^\top \Lambda^{1/2}
\end{equation}

Finally, if $M$ is equal to the covariance matrix $M^{(\alpha)}_\epsilon$ introduced in Section~\ref{sec:kernels}, then~\eqref{eq:relation-eigenfunctions} allows us to write the diffusion map coordinates as
\begin{equation*}
  \lambda_n \psi_n = \sqrt{\lambda_n} \phi_n.
\end{equation*}
Taking the expectation value of~\eqref{eq:f-tilde} and letting $f_n = \sqrt{\lambda_n} \, \E[ \tildexi_n ] \in \R$ in
\begin{equation*}
  \E[ \tf(x) ]
  =
  \sum_{n \in \mathbb{N}} \sqrt{\lambda_n} \, \E[ \tildexi_n ] \, \phi_n(x),
\end{equation*}
we conclude that
\begin{equation}
  \label{eq:nystrom-expectation}
  \E[ \tf(x) ]
  =
  \sum_{n \in \mathbb{N}} f_n \, \psi_n(x)
\end{equation}
is the Nystr\"om extension of the deterministic function $\E[ \tilde{f} ]$ in terms of the diffusion map coordinates corresponding to the operator corresponding to the kernel $p^{(\alpha)}_\epsilon$.
In other words~\eqref{eq:nystrom-expectation} is a particular case of geometric harmonics approximation.


\section{Uncertainty in GP vs Error in GH}\label{sec:uncertainty}

The correspondence between Gaussian Processes and Geometric Harmonics goes beyond function regression. Both concepts involve theory about the behavior of the surrogate function away from the given data points. In Gaussian Processes, probability theory dictates uncertainty bounds through the standard deviation of the normal distribution in the function space. In Geometric Harmonics, the uncertainty of the function value away from given data is provided through limits on the number of eigenvalues and corresponding number of eigenfunctions used to accurately approximate the function at a certain distance from the data.

In Gaussian Processes, the uncertainty of the process at a new point $x^*$ is given by~\cite{rasmussen-2005}
\begin{equation}\label{eq:gp uncertainty covariance}
 \Sigma_*=k(x^*,x^*)-k(x^*,X)\left[k(X,X)+\sigma^2I\right]^{-1}k(X,x^*)\approx \text{cov}(x^*).
\end{equation}
This formula defines the conditional probability of $f$ at $x^*$ given the data $\{ x_1, x_2 \dotsc \}$ collected as column vectors in the matrix $X = [x_1, x_2, \dotsc]$.
Crucially, the covariance does not involve the actual function values at the sample points $x$: it is only the location of the points that matters.

In turn, for Geometric Harmonics, the uncertainty of the function values for new data points is treated as an extension problem away from the given data points. As, ultimately, the function to extend is written as a linear combination of eigenfunctions of the Laplace-Beltrami operator, the extension to new data points involves {\em extending those eigenfunctions} away from the given data (manifold) into the ambient Euclidean space. Extendability is defined by Lafon through definition~\ref{eq:definition extendability}.
\begin{definition}\label{eq:definition extendability}
A function $f$ defined on $\Gamma$ is said to be $(\eta,\delta)$-extendable if for a given couple $(\eta,\delta)$ of positive numbers
\begin{equation}
\sum_{j\in S_\delta}|\left\langle\psi_j,f\right\rangle_{\Gamma}|\geq (1-\eta)\|f\|^2_{\Gamma}.
\end{equation}
\end{definition}
Here, $S_\delta = \left\lbrace j \text{ such that } \lambda_j>\delta \lambda_0\right\rbrace$.
As a consequence also noted in~\cite{coifman_geometric_2006}, if $j\in S_\delta$, then the eigenfunction $\psi_j$ itself is $(\eta,\delta)$-extendable for all $\eta>0$.

Instead of using this concept of extendability, we reinterpret the estimated error of Geometric Harmonics so as to more closely correspond to the uncertainty of Gaussian Processes. This will lead to an exact similarity of the two approaches (under certain assumptions), and also allows us to interpret uncertainty of Gaussian Processes in a different, hopefully informative, way.
We first interpret equation~\eqref{eq:gp uncertainty covariance} in a Geometric Harmonics context. The first term, $k(x^*,x^*)$, is the evaluation of the kernel function at a new point. The second term, $k(x^*,X)\left[k(X,X)+\sigma^2I\right]^{-1}k(X,x^*)$, is an approximation of the first term, using a Gaussian Process with kernel values at known points $X$. We now perform this approximation with Geometric Harmonics instead, using equation~\eqref{eq:gh reg} to predict the function $k(x^*,\cdot)$ at $x^*$ using the function values $k(x^*,X)$ at the known data points $X$:
\begin{equation}
\hat{k}(x^*,x^*)=k(x^*,X)^T V\Lambda^{\dagger} V^T k(x^*,X).
\end{equation}
Then, exactly as in equation~\eqref{eq:gp uncertainty covariance}, we just subtract the approximation of the kernel value at $x^*$ from the true value, to obtain the error estimate:
\begin{equation}\label{eq:gh error}
\Sigma_*^{GH}:=k(x^*,x^*)-\hat{k}(x^*,x^*).
\end{equation}

\begin{figure}[ht]
\includegraphics[width=1\textwidth]{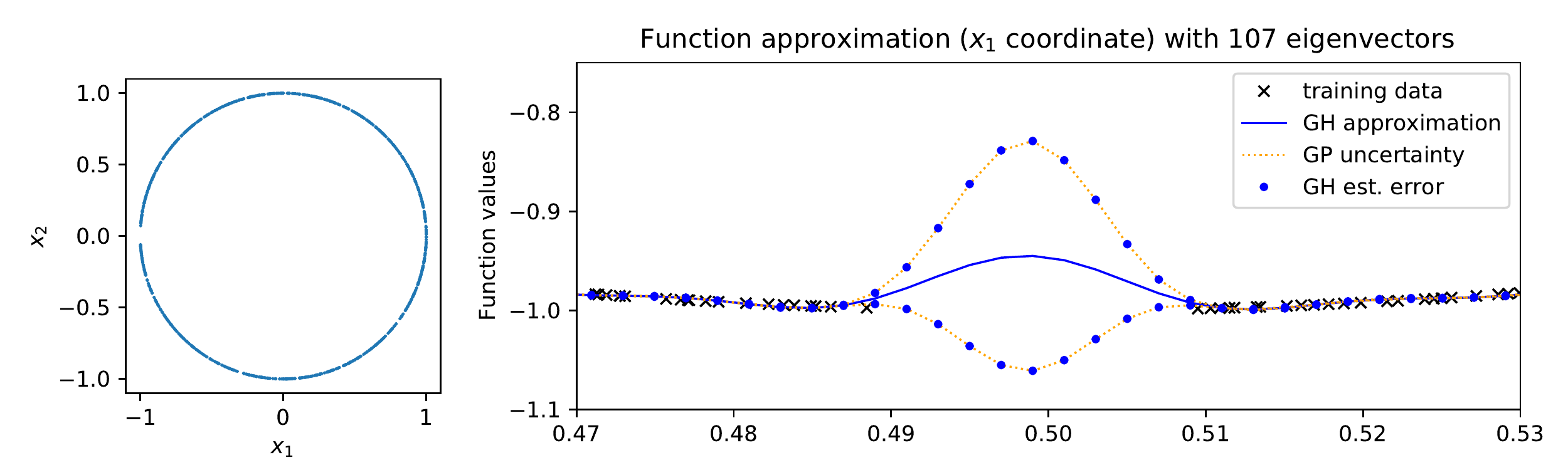}
\caption{\label{fig:gp uncertainty vs gh error}Left: Dataset with a small gap on the left. Right: Estimated error of Geometric Harmonics compared to estimated uncertainty of a Gaussian Process inside the gap.}
\end{figure}

As a small demonstration, we estimate error with Geometric Harmonics as evaluated by equation~\eqref{eq:gh error} for the dataset shown in figure~\ref{fig:gp uncertainty vs gh error} and compare it to the uncertainty estimated by a Gaussian Process. As expected, the error increases in the region where there are no data points, and reaches the maximum of $k(x^*,x^*)$ in the center of the ``missing interval", at $(-1,0)$. The error and uncertainty estimates agree.

\begin{figure}[ht]
\includegraphics[width=1\textwidth]{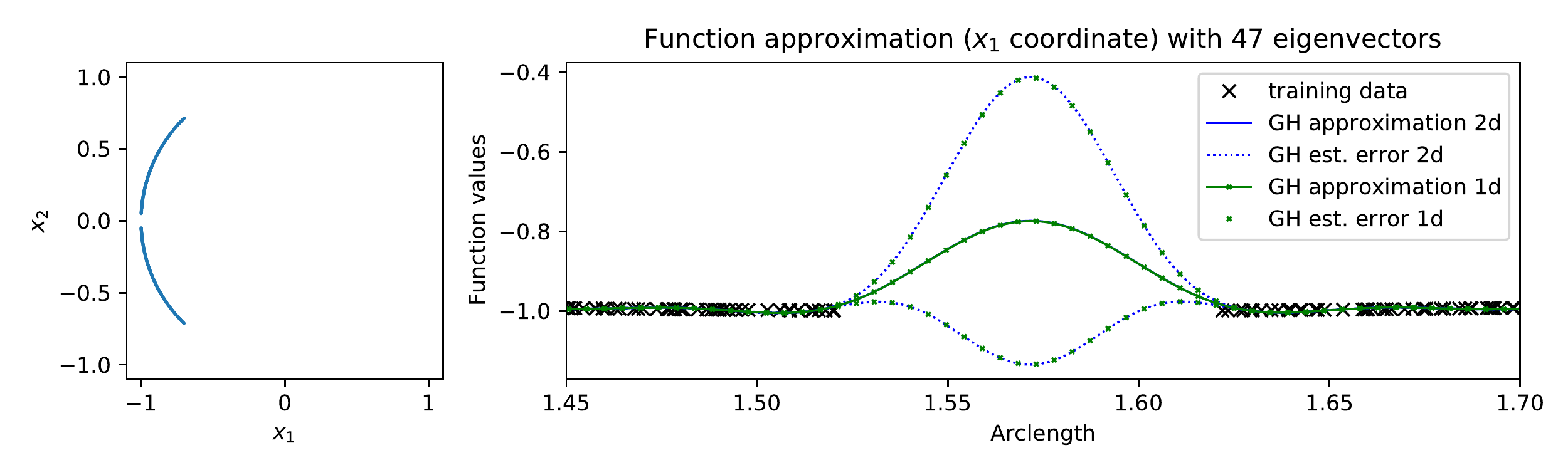}
\caption{\label{fig:two_d_uncertainty}Left: portion of the dataset used to evaluate full space vs. latent space error estimations. Right: Error estimated with 2d data compared to estimation with isometric, 1d arclength embedding.}
\end{figure}

In figure~\ref{fig:two_d_uncertainty}, we demonstrate that the estimation of the error is independent of the ambient space, if the embedding into the latent space is isometric and we use a kernel that only depends on the distance. We chose a subset of the points used for figure~\ref{fig:gp uncertainty vs gh error} so that we can embed them in one dimensions using the arclength along the circle. 

\subsection{Interlude: Inductive Biases}\label{sec:inductive biases}

Choosing a different kernel amounts to a different inductive bias for each of our two kernel methods. In particular, for Gaussian Processes, different kernels imply different prior distributions in the space of functions, which in turn leads to different generalization behavior. This is well known~\cite{rasmussen-2005}, and for certain kernels, the exact types of functions that can be approximated and extended to unseen data can be/have been described.

What is discussed less often is how given kernels actually extend functions away from the given data---that is, how they generalize into the ambient space. In the thesis of Lafon~\cite{lafon-2004}, the author outlines how choosing a squared exponential kernel implies that the surrogate function has zero gradient in the normal direction (in the ambient space) of the manifold on which the data points lie.

This implies that functions evaluated {\em on the data} are extended \textit{in a constant way} away from the data, as long as the new data point is close enough to the original (before ``caustics" arise). The value of this constant, as expected, depends on the function value(s) at the data point(s) closest to the new  point.

\subsection{A surprising example for inductive bias}
\begin{figure}
    \centering
    \includegraphics[width=.5\textwidth]{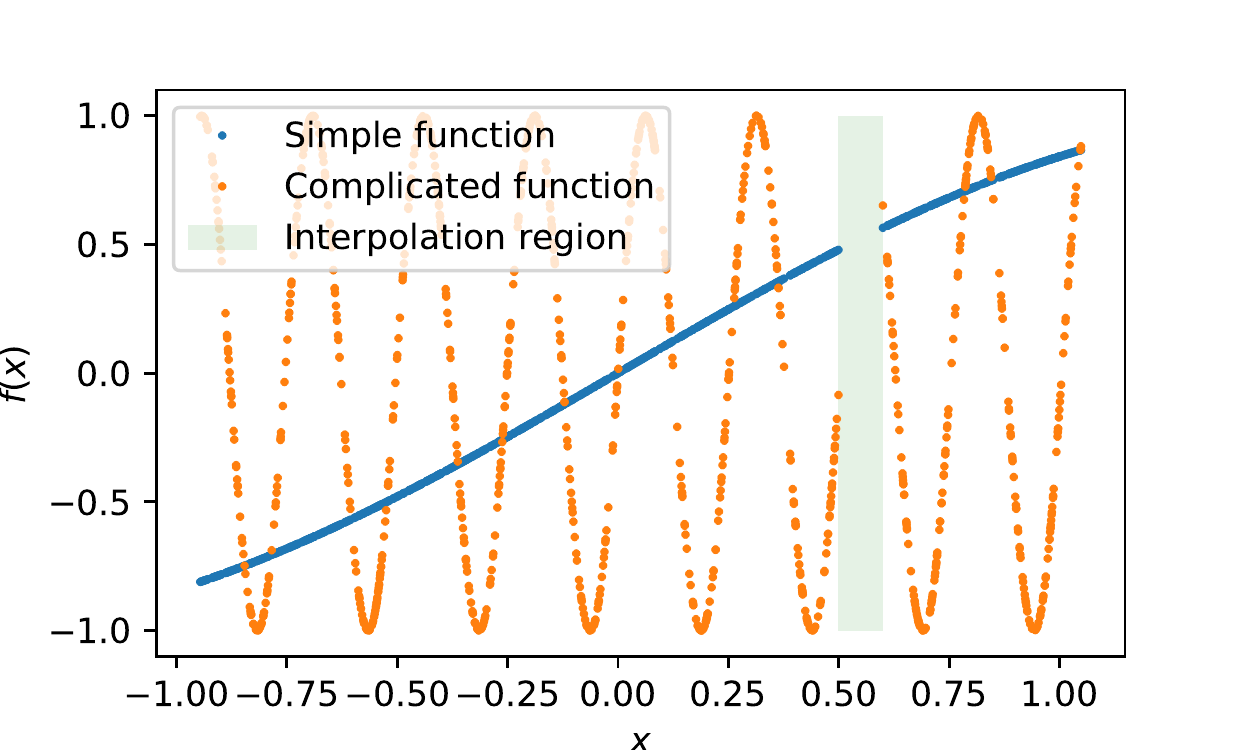}
    \caption{Dataset for the example to demonstrate inductive bias for GP uncertainty. In the assessment of the bias, we focus on the interpolation region where no data is available.}
    \label{fig:gp inductive bias dataset}
\end{figure}

\begin{figure}
    \centering
    \includegraphics[width=1\textwidth]{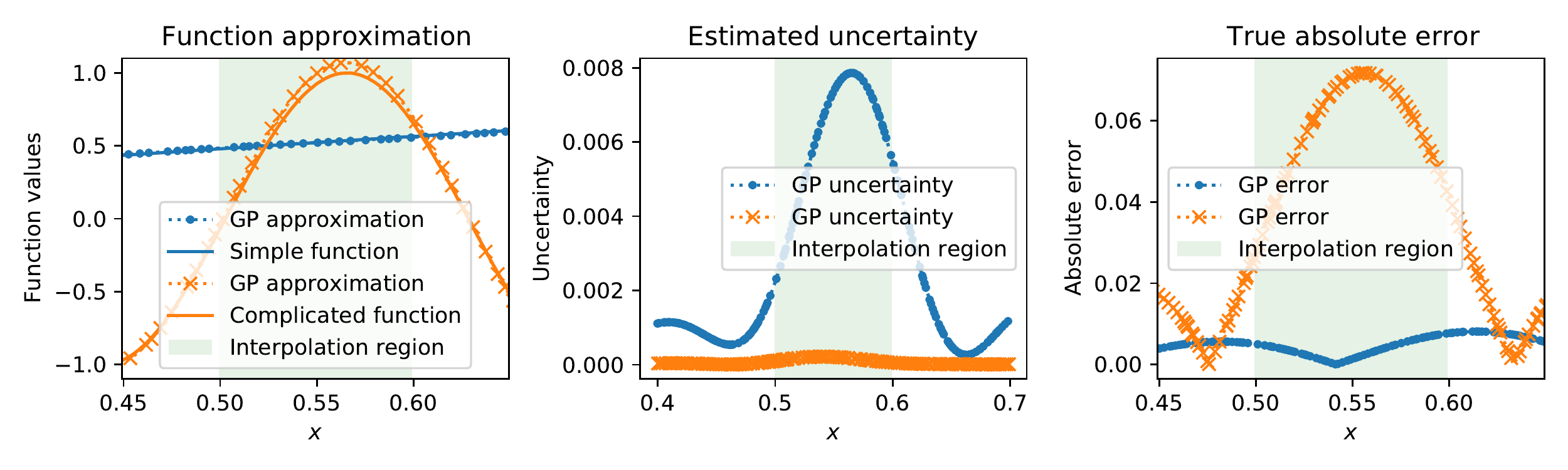}
    \caption{Inductive bias of GP uncertainty estimation. The true error in the interpolation region is smaller for the simple function, but the estimated uncertainty is higher. This is counter-intuitive, but can easily be explained through the number of eigenvectors chosen (implicitly) by the kernel scale hyper-parameter tuning of the parameter $\alpha$. Of course, the same inductive bias is present for Geometric Harmonics, albeit more explicit (because the number of eigenvectors is explicitly chosen).}
    \label{fig:gp inductive bias error}
\end{figure}

We now briefly outline an---at first sight---surprising result of the analysis of inductive bias of the uncertainty estimates of Gaussian Processes, using their correspondence to Geometric Harmonics.
For the sake of simplicity, we assume we want to employ a Gaussian Process
in just one dimension, to interpolate two functions, one ``simple'' and one ``complicated'' in terms of their Dirichlet energy, between two, one-dimensional segments of data (see Figure~\ref{fig:gp inductive bias dataset}). To assess the quality of the interpolation, we use the estimated GP uncertainty in the region of interpolation.
We then change the regularization parameter $\alpha$ in GP regression (see Equation~\eqref{eq:gp reg}) to best fit the data, and plot the estimated uncertainty {\em as well as the actual error} of the estimated mean versus the true function inside the data gap (Figure~\ref{fig:gp inductive bias error}).
The error for the more complicated function is higher, as expected. However, the estimated uncertainty for it is much lower than the one for the simple function.

This may be a surprising result for GP regression, but can be easily explained through the correspondence to GH. Changing the parameter $\alpha$ directly influences how many eigenfunctions of the kernel are considered---{\em implicitly} for GP, because of the stronger regularization, and  {\em explicitly} for GH, because it selects the floor for the eigenvalues retained. 
For the ``simple'' function, a few eigenfunctions suffice for accurate  approximation. However, the GP uncertainty is being computed by approximation of the \textit{kernel function}, not the empirical function that is being approximated. Now, the kernel is the same for both our complicated and our simple function; yet, the number of eigenfunctions retained is lower for the simple function. As a result, the error in the kernel function approximation is larger for the simple function case.

\section{Numerical algorithms for GH and GP}\label{sec:numerical algorithms}
We now offer another perspective on the relationship between Geometric Harmonics (GH) and Gaussian Process Regression (GPR).
This time we view both methods through the lens of spectral filtering~\cite{hansen_discrete_2010}.

GH expresses\todo{The notation here is confusing. GHs are expectation values, so we should write that unambiguously in~\eqref{eq:gh reg} and~\eqref{eq:gp reg}.} an approximation $\mu^{GH}_\star$ to a function $f$ at new points $x^\star$ through
\begin{equation}\label{eq:gh reg}
  \mu^{GH}_\star=\underbrace{k(x^\star,X) V\Sigma^{\dagger}}_{{\Phi}(x^\star)} \underbrace{V^T f(X)}_{\left\langle \phi,f(X)\right\rangle}
  =
  V \Sigma V^T,
\end{equation}
while in GPR, $\mu^{GP}_\star$ is the expected value of a GP at $x^\star$ expressed through\todo{Do we explain what is meant by $X^\prime$ elsewhere?}
\begin{equation}
  \label{eq:gp reg}
  \mu^{GP}_\star
  =
  k(x^\star, X) (k(X,X^\prime) + \alpha^2 I)^{-1} f(X),
  \quad
  \alpha > 0.
\end{equation}
Both~\eqref{eq:gh reg} and~\eqref{eq:gp reg} are regularized solutions to the linear system
\begin{equation}
  \label{eq:linear problem}
  A y = b,
\end{equation}
where $A = k(X, X^\prime)$ is a symmetric and positive definite matrix and $b = f(X)$.
After the solution $y$ is found, evaluation at new points $x^\star$ is reduced to a linear combination of kernel functions,
\begin{equation}\label{eq:gp and gh extension to xstar}
\mu_\star=k(x^\star,X)y=\sum_{i=1}^N k(x^\star,X_i)y_i.
\end{equation}
We drop the superscripts to $\mu_\star$ because \eqref{eq:gp and gh extension to xstar} is the same formula for GPR and GH.
In general, if $A$ in~\eqref{eq:linear problem} is singular or ill-conditioned, then the solution to its corresponding linear system can be obtained via Tikhonov regularization
as
\begin{equation}
  \label{eq:tikhonov regularization problem}
  y = \argmin_{z \in \R^N} \|A z - b\|_2^2 + \| \Gamma z \|_2^2,
\end{equation}
where $\Gamma$ is the Tikhonov matrix, to be discussed in what follows.
Clearly, the minimizer $y$ of~\eqref{eq:tikhonov regularization problem} depends on the choice of $\Gamma$.
The solution $y$ of~\eqref{eq:tikhonov regularization problem} follows immediately by setting the gradient of $\|A z - b\|_2^2 + \| \Gamma z \|_2$ to zero, and is explicitly given by
\begin{equation}
  \label{eq:tikonov solution}
  y
  =
  (A^T A + \Gamma^T\Gamma)^{-1} A^T b.
\end{equation}
\noindent
GH and GPR differ by the choice of $\Gamma$.
For GH, in~\eqref{eq:gh reg}, the solution to~\eqref{eq:linear problem} is regularized through the Moore-Penrose pseudo-inverse of $A$. This inverse is related to (\ref{eq:tikonov solution}) by choosing $\Gamma=\sqrt{\delta} I$ and defining $\hat A=V\Sigma_\alpha V^T$ where $\Sigma_\alpha$ is the diagonal matrix with entries
\begin{equation*}
  \left( \Sigma_\alpha \right)_{ii}
  =
  \begin{cases}
    \sigma_i, & \text{if $\sigma_i < \alpha$,} \\
    0, & \text{otherwise}.
  \end{cases}
\end{equation*}
Consequently,
\begin{equation}
  A^\dagger
  =
  \lim_{\delta\to 0^+} (\hat{A}^T \hat{A} + \delta I)^{-1} \hat{A}^T
  =
  V \Sigma^\dagger V^T.
\end{equation}
For GPR, the choice $\Gamma=n^{1/2}\alpha A^{1/2}$ is employed to solve~(\ref{eq:tikhonov regularization problem}), where $A^{1/2}$ is the unique Cholesky factor of $A$.
This turns~\eqref{eq:tikhonov regularization problem} into
\begin{equation}
  \min_{z \in \R^N} \| A z - b \|_2^2 + n \alpha^2 z^T A z,
\end{equation}
with the solution typically employed in ridge regression,
\begin{equation}
  y
  =
  (A^T A + \Gamma^T \Gamma)^{-1} A^T b
  =
  (A + n \alpha^2 I)^{-1} b.
\end{equation}

\begin{figure}[ht]
  \centering
  \includegraphics[width=0.666\columnwidth,trim=0.25cm 3cm 0.25cm 4.125cm,clip]{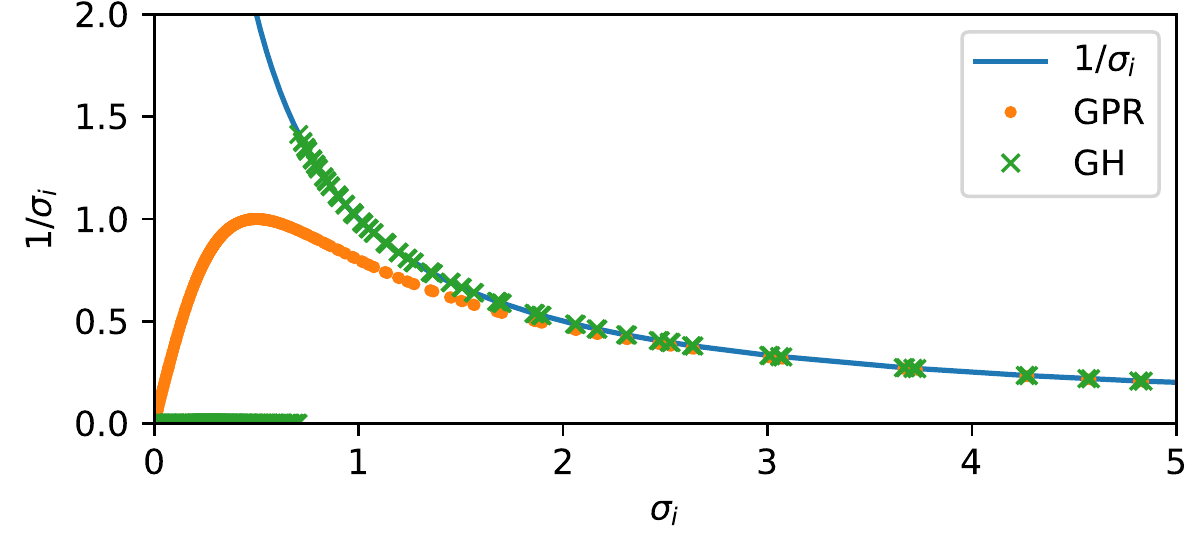}
  \caption{\label{fig:GH_GPR_singularvalues}Example of the different treatment of the inversion of singular values between GPR and GH (computed for a squared exponential kernel on the unit circle).}
\end{figure}

Since $A$ is a symmetric and positive definite matrix, we can write $A = V \Sigma V^T$, and then the solution to the GPR regularization problem can also be written as
\begin{equation}
  y
  =
  V D V^T b,
  \quad \text{with} \quad
  D_{ii}
  =
  \frac{\sigma_i^2}{\sigma_i^2 + \alpha^2},
\end{equation}
which makes the difference in the regularization of the singular values compared to GH apparent:
In GPR, $\sigma_i<\alpha$ results in $D_{ii}$ decaying to zero, while in GH, the values $D_{ii}$ are directly set to zero for $\sigma_i<\alpha$. This means that for $\sigma_i\to 0$, the components of the vector $b$ in the direction of eigenvectors $v_i$ associated to $\sigma_i$ are being increasingly suppressed by GPR, while they are completely removed in GH (see Fig.~\ref{fig:GH_GPR_singularvalues}).


\section{Dimension reduction}\label{sec:dimension reduction}

In this section, we discuss the relation of the Gaussian Process and Geometric Harmonics frameworks to dimension reduction methods.
Strictly speaking, dimension reduction is not directly related to function approximation, because the function values we try to approximate are typically not related to the ambient space in which the manifold on which the data points lie (the ``data manifold'') is embedded.
However, if we employ kernel methods, the kernel function itself is defined on the whole ambient space of the data, not just the input manifold. This implies we actually approximate the function on the full ambient space, even though the lower intrinsic dimensionality of the data suggests that we could work on a reduced space.
Motivating examples where the intrinsic dimension can be much smaller than the ambient dimension (as we mentioned in the introduction) are singularly perturbed dynamical systems or certain types of multi-scale optimization problems, as mentioned at the introduction (see also \cite{Pozharskiy-2020}, and in the context of neural network training, see~\cite{Wang-2020}).

\subsection{Assumptions on the data and process}

For the remainder of this section, we assume that the data points available to us lie on a low-dimensional manifold, and are embedded in a higher-dimensional Euclidean space.
In classical GP regression, this assumption would not be exploited, and the process would be constructed on the full ambient space.
Here we advocate exploiting the low dimension of the manifold by first embedding the data points in a lower dimensional ambient space, and then performing the function approximation steps there (and iterating the process as necessary).

This approach is useful for both GP and GH, because the computational complexity of distance computations usually depends on the ambient space dimension. 
In addition, in a space with lower ambient space dimension, hyper-parameter tuning, training, and inference are all sped up.

If the kernel used in GP or GH only depends on the distance between data points, and the embedding into the lower-dimensional space is isometric (preserves distances), then the expected value in GP and the function approximation in GH are both identical to the full space results.

If either the kernel depends on more than just the distance, or if the embedding is not isometric, the approximation quality will be different (not necessarily worse!). The same is true for the GP uncertainty (or GH error) estimates.

Constructing a lower-dimensional embedding space has immediate benefits for Bayesian optimization, where a Gaussian Process surrogate model for the original function is already being constructed. The idea of using eigenfunctions of the process to embed available data can then hopefully speed up the optimization.

However, the mapping to the new embedding space has a new complication, that is not present in the standard approach. If we compute a new candidate point to evaluate the original function in the new embedding space, we cannot immediately evaluate the original function on it---we must first map back to the original, high-dimensional space.
This is the problem of ``lifting'' in multiscale computations~\cite{theodoropoulos-2000,kevrekidis-2003,kevrekidis-2009,Arbabi-2020,Vlachas-2020},
the same problem that generative models from deep learning also address~\cite{goodfellow-2014, noe-2013,noe-2018,Chen-2018b}.

It is possible to write the original, ambient space coordinates of the points {\em on the reduced manifold} as functions {\em on this manifold}, i.e. as functions of new, low-dimensional embedding of the manifold in terms of the available leading kernel eigenfunctions.
Approximating these functions with either GP or GH endows the approximations with their corresponding uncertainty/error estimates.

We can then use these approximations as ``lifting coordinates'' and obtain the desired (possibly expensive to evaluate) function in the high dimensional ambient space. We can then immediately, ``streamingly'', use this new data point to update our eigenfunction estimates we use as low-dimensional manifold embedding coordinates. 
More practically (and in the same spirit of using the same preconditioning for a few iterations before updating it) we may perform several algorithm iteration steps (e.g. several BO steps) before updating our kernel matrices and the corresponding low dimensional embeddings. The optimal number of ``iterations before updating'' now becomes a question for the numerical analysis of the overarching iterative algorithm.

\subsection{A dimension reduction algorithm related to GH and GP: Diffusion Maps}

Diffusion Maps is an algorithm that is directly related to the concept of Geometric Harmonics, but is exploiting the fact that, in general, sufficiently smooth low-dimensional manifolds can in principle be embedded in just a few eigenfunctions of their Laplace-Beltrami operator---i.e., exactly the Geometric Harmonics. Instead of computing several hundreds of these functions so as to then be able to represent functions as linear combinations of them, we only compute the leading {\em independent} eigenfunctions~\cite{Dsilva-2018,chen-2019} related to the smaller eigenvalues of the operator, and then use them as new coordinates of the manifold. 

{
An issue specific to Diffusion Maps is that---due to zero Neumann boundary conditions implicit in the standard algorithm---the eigenfunctions of the Laplace-Beltrami operator all have vanishing gradients on the boundary of the data manifold. The gradient will decay quickly with decreasing kernel bandwidth, and will be exactly zero in the limit. As we explicitly want to extend beyond the boundary of a small sample of points, embedding into eigenfunctions with vanishing gradients is an issue: The objective function in the new embedding space will have an infinite gradient exactly at the point where we want to extend it.
In our toy example, we use a relatively large kernel bandwidth due to the small sample size, which alleviates the issue.
Several techniques already exist to completely alleviate it systematically, including Sine Diffusion Maps~\cite{georgiou-2017}, Diffusion Maps for Embedded Manifolds with Boundary~\cite{Vaughn-2019}, as well as Ghost Point Diffusion Maps~\cite{Jiang-2020}.
}

\subsection{Uncertainty in latent spaces}

\begin{figure}[ht]
    \centering
    \includegraphics[width=0.85\columnwidth]{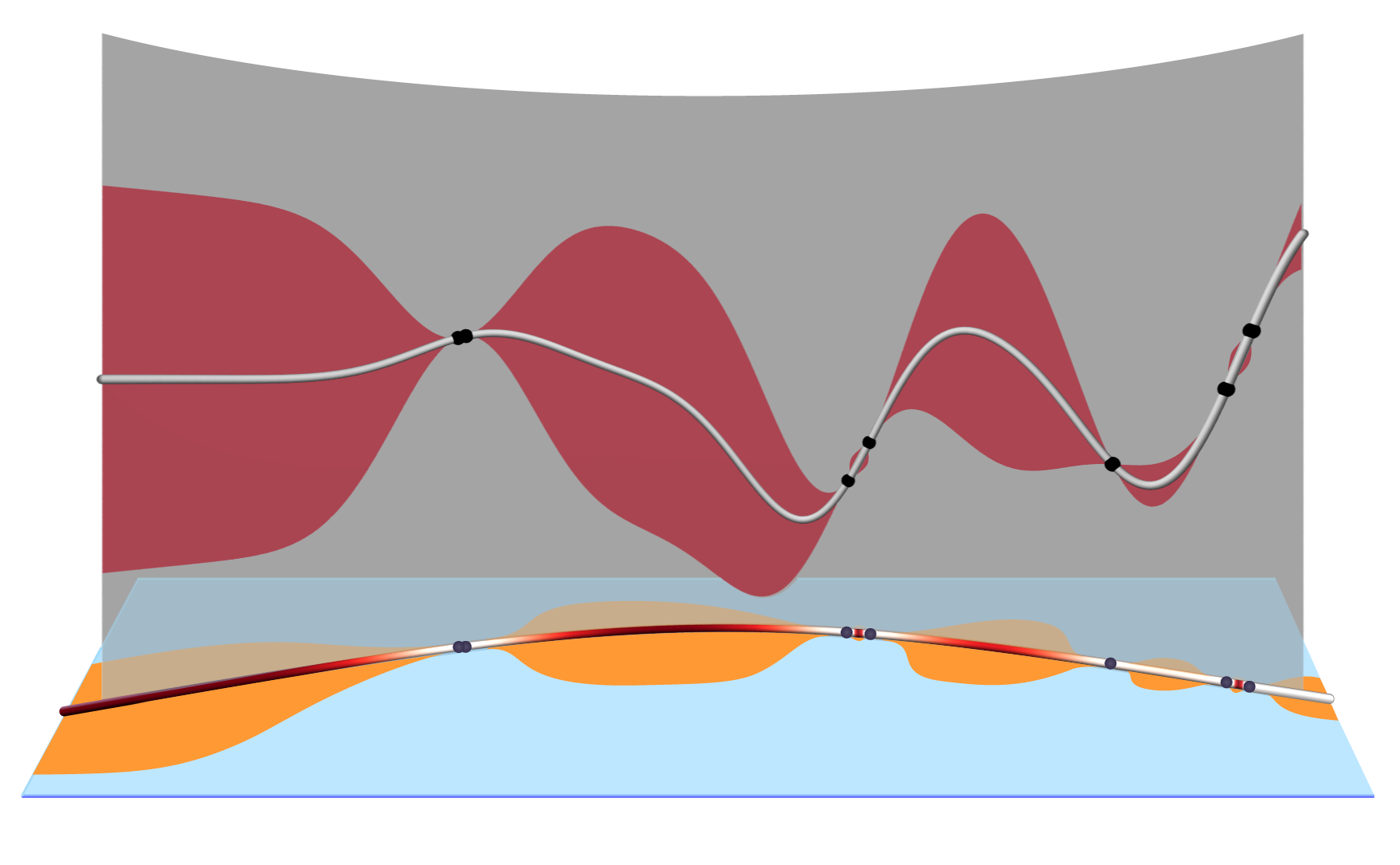}
    \caption{Gaussian Process defined over a reduced, one-dimensional domain:  a curve embedded in a 2D space (bottom).
    A few samples of the function of interest \textbf{(black points)} lead us to a surrogate model: the expectation is the gray line, and its uncertainty over the 1-D curved manifold is plotted \textbf{(red)} on the vertical gray ``curtain". The uncertainty (in the two dimensions) of lifted points on this one-dimensional reduced manifold is shown in orange at the base of the figure.}
  \label{fig:gp-curved-domain-illustration}
\end{figure}

Figure~\ref{fig:gp-curved-domain-illustration} depicts the aforementioned approach: the original objective function is defined in some ambient space (here, two-dimensional) but there is, nevertheless, a one-dimensional manifold (curve) outside of which the objective grows quickly.
Thus, it makes sense to conduct our optimization process as if the objective function was defined solely on the 1D manifold.

\begin{figure}[ht]
  \centering
  \includegraphics[width=0.85\columnwidth]{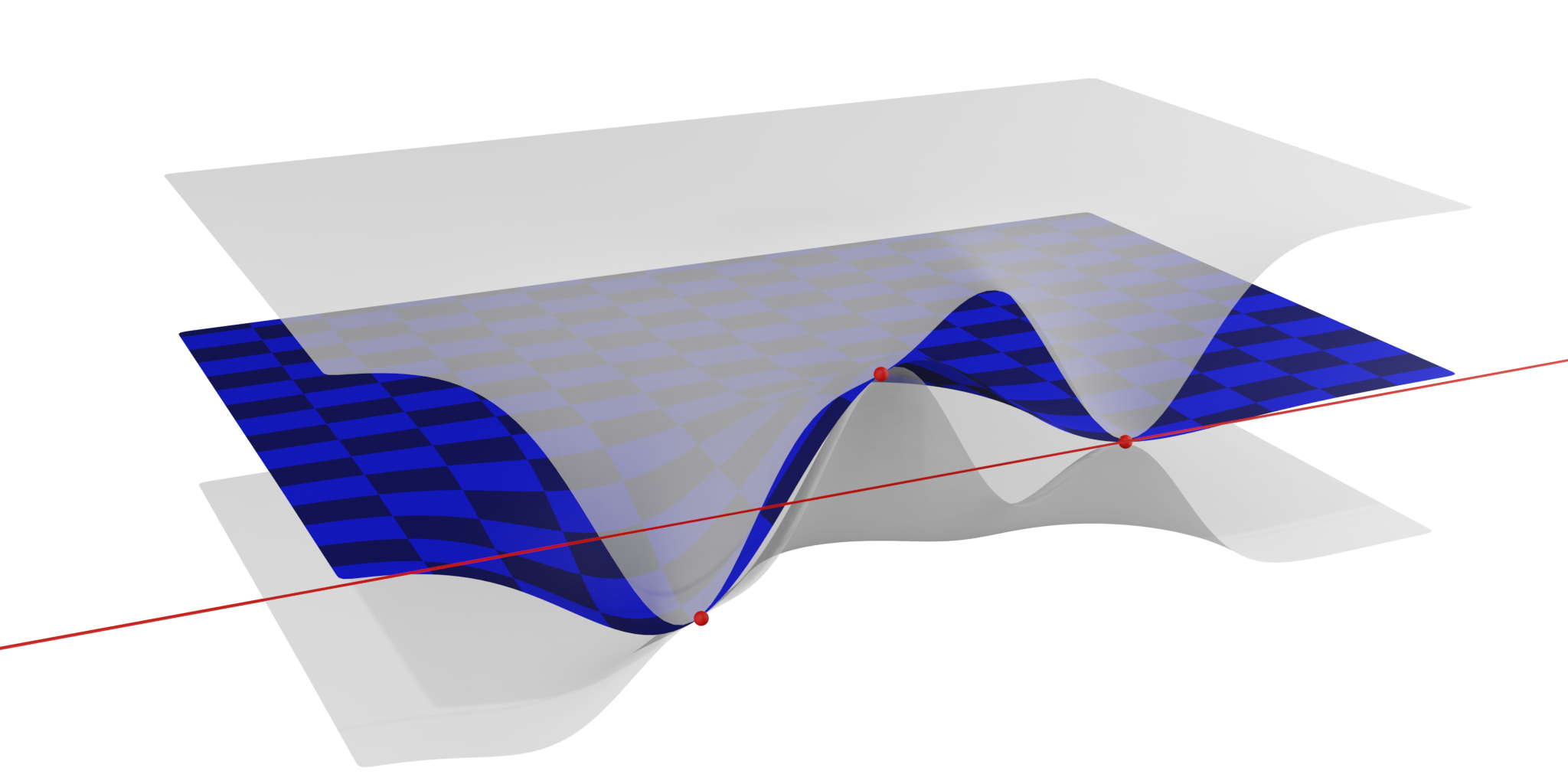}
  \caption{Representation of the uncertainty of a Gaussian process regressor for a function of two variables when the data (red spheres) lies on a one-dimensional manifold $M$ (red line).
    What  we  see  in  the  three-dimensional  Euclidean space is the graph of the mean, $\{ (x_1, x_2, y) \in \R^3 \mid y = f(x_1, x_2) \}$, of the GP regressor when it is evaluated off manifold.
    The checkered sheet over the entire embedding space indicates the expected value; the two gray sheets above and below indicate the uncertainty ---they are separated by the standard deviation of the GP. The asymptotic distance is equal to $\lim_{\| y \| \to +\infty} k(x, y)$, which depends on $k$, the kernel of choice, with $x \in M$ and $y \in \R^2$.}
  \label{fig:gp-2d-3d-uncertainty}
\end{figure}

Figure~\ref{fig:gp-2d-3d-uncertainty} illustrates the behavior of a Gaussian process regressor for a two-variable function determined by points sampled from a one-dimensional manifold (a line) embedded in a two-dimensional ambient space (horizontal plane).
The vertical dimension represents the value of the regressor as a function of the ambient space coordinates.


\section{Discussion and outlook}\label{sec:discussion}

We briefly reiterate the issues that have been discussed: Comparing equations (\ref{eq:gh reg}) and (\ref{eq:gp reg}) makes the similarity of the  \textit{Gaussian Process} and \textit{Geometric Harmonics} approaches apparent.
It also points out the first difference: inverting the kernel matrix for Gaussian Process Regression uses regularization through $\sigma$, whereas Geometric Harmonics employ the pseudo-inverse.
This could be explained by the conceptual background of the two approaches, one rooted in statistics (where noise is ubiquitous), the other in harmonic analysis (where non-invertibility would perhaps be the result of numerical inaccuracies but not stochastic effects).

These conceptual differences lead to discrepancies in the interpretation of the results, as well as differences in the augmentation of the core method with additional information. 
For example, in Gaussian Process Regression it is customary to include the standard deviation of the posterior distributions of ${f}_{x^*}$, which can be computed for any point $x^*\in\mathcal{M}$ with a simple formula (see $\Sigma_*$ in equation~(\ref{eq:gp uncertainty covariance})).
For Geometric Harmonics, there exists the concept of multiscale analysis~\cite{coifman-2005b} with well-founded theory. 
A marked trend in the community surrounding Gaussian Processes is to approach multiple scales from a more engineering-driven perspective, in the form of ``deep Gaussian Processes'' (in various forms, e.g. \cite{lee-2017b,pang-2018,cutajar-2018}).
Some ad-hoc approaches to multi-scale GPR exist, see e.g.~\cite{denzel-2018b} (in an update scheme, the total number of computed points is divided into small subsets and a GP is built for every subset, with an ordering such that the GP with earlier data serves as prior for the next GP).
An interesting result for Gaussian Process Regression concerning artificial neural nets is that infinitely wide neural nets with a smooth nonlinearity are identical to a Gaussian Process with a kernel adapted to the specific network nonlinearity~\cite{neal-1996,lee-2017c}. Combining this result with the tools of harmonic analysis used in the multi-scale approach of Geometric Harmonics is then a compelling direction.

In this work, we  have established the formal correspondence between GP and GH based on samples on a low-dimensional manifold embedded in a (higher-dimensional) ambient space. 
We have also pointed out the similarities between error estimates in GH and uncertainty estimates in GP.
We also noted in passing certain dissimilarities: the differences in the corresponding numerical implementations, as well as a (simple, but surprising to us) difference in the corresponding inductive biases.

Beyond establishing the formal correspondence, an important goal is to exploit it, in hopefully accelerating numerical algorithms, like BO, by taking advantage of the possible {\em low-dimensionality of the algorithm iterates themselves}. If this is the case and if the iterative algorithm itself quickly evolves to a lower dimensional manifold, then the conditional KL expansion underlying both methods can potentially lead to reduced computational cost and number of iterations. This is where the low-dimensional embedding features of GH may assist the regression performed by GP.

There are many clear directions for making these observations useful in both numerical analysis and scientific computation.
The focus is on iterative algorithms in multiscale contexts---whether they relate to simulation, or fixed point/parametric analysis, or optimization and control.
The main idea is that the iterates of the algorithm, in the high-dimensional ambient space naturally ``concentrate'' (due to scale separation of time scales, eigenvalues, sensitivities) on lower-dimensional manifolds (e.g. see \cite{dietrich-2020b}).
The common language of conditional KL expansions can then seamlessly (a) deduce the parametrization of a useful reduced ``latent space''; (b) provide a local surrogate (what we would like to call a ``targeted surrogate'' in the sense that it is not global, but rather ``just enough" for the next algorithm iteration) in this latent space that (c) can be used to design the next algorithm iteration in this targeted latent space; and then also (d) translate the results to the full space (``lifting''), where the full model will be briefly used (to briefly simulate, or to evaluate the expensive objective function).

This progression (obtaining high dimensional data, analyzing them to find a useful latent space, with a local ``targeted surrogate" that will help design the next expensive high dimensional experiment/function evaluation) is a natural progression in multiscale numerics (such as the variable free/equation free computations previously proposed).

Usefully going back and forth between low-dimensional and high-dimensional description of the same problem is the linchpin of model reduction. There is a prevailing perception today that one first learns a global accurate surrogate model, and only then uses it to perform reduced numerical tasks. It seems clear to us that the critically important low-dimensional surrogate model {\em does not need to be global} in order to design the next high-dimensional experiment/function evaluation.
The conditional KL expansion provides a natural setting in which both the data reduction and ``targeted surrogate'' in the reduced space are constructed and streamingly updated in a continuous dialogue with the (fine scale, expensive) full model.

Careful numerical analysis of all these reasonable extensions and options (e.g. the importance and benefits of streaming) will help decide which of them may benefit different types of problems. 

{\bf Acknowledgements}  This work was partially supported by the DARPA ATLAS program (Dr. J. Zhou) and the US Department of Energy.


\bibliographystyle{elsarticle-num-names} 
\bibliography{lit.bib}

\appendix

\counterwithin{figure}{section}

\section{Towards reduced Bayesian Optimization}
\label{sec:towards-reduced-bayesian-optimization}



\begin{figure}[ht]
  \includegraphics[width=0.9\columnwidth]{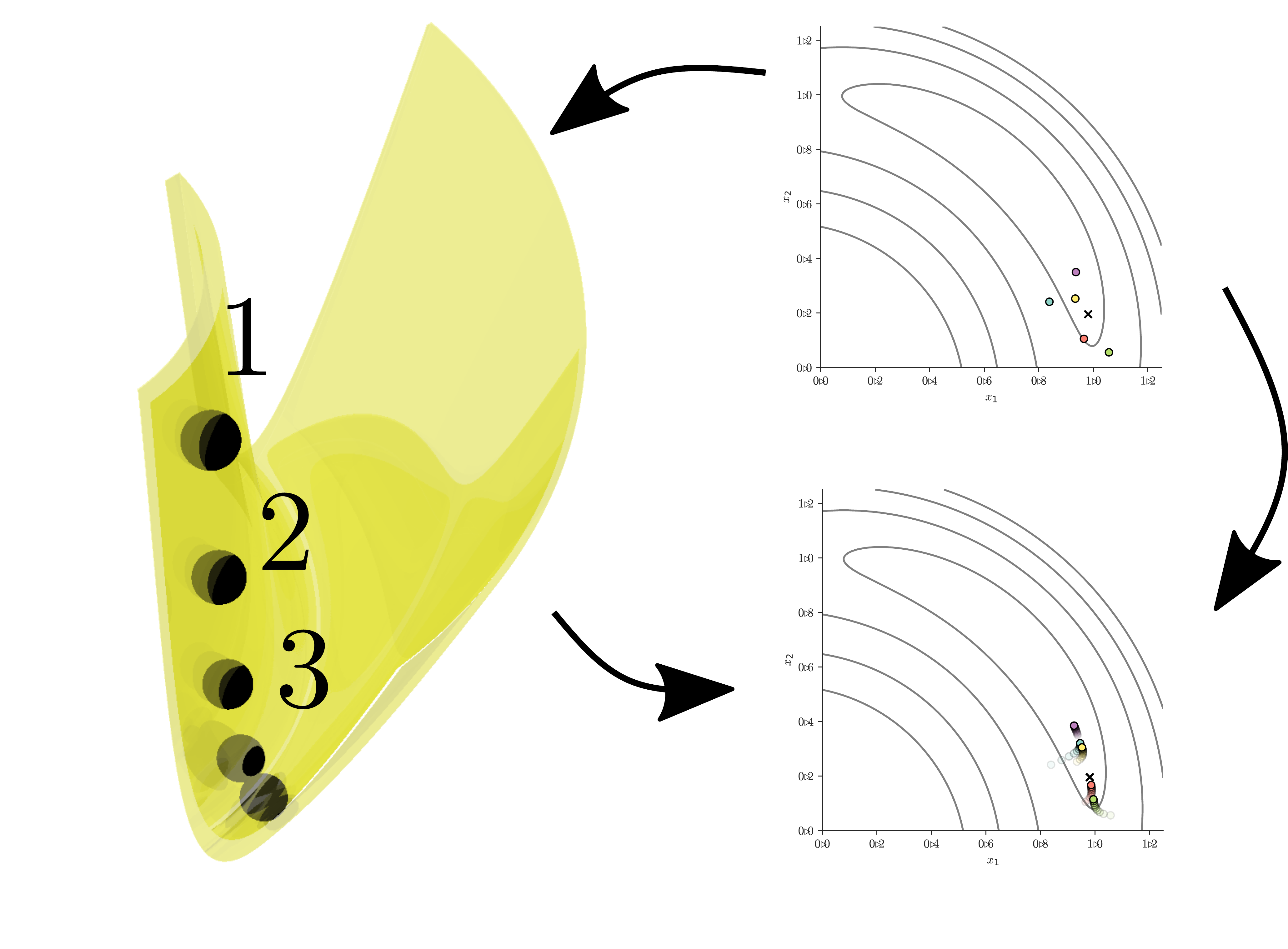}
  \caption{We start our optimization process by relaxing a number of randomly-drawn samples ($\bullet$) around an initial point ($\times$) on the canyon-shaped objective function (represented in yellow on the left and by the gray contours on the other two figures). This is accomplished by running a few steps of steepest descent, quickly relaxing the initial sample points on the two-dimensional canyon walls (upper right) down to the (effectively one-dimensional) canyon bottom (lower right).}
  \label{fig:relaxation-canyon}
\end{figure}

Here we discuss a simple minimization problem used to produce elements of Figure~\ref{fig:canyon-artistic} in \secref{sec:introduction} to motivate combining Bayesian optimization (BO) with dimensionality reduction in multiscale optimization problems with ``canyon-like" objective functions. In such problems we expect that a few initial steps of gradient descent  suffice to relax each of the sampled points (see Figure~\ref{fig:relaxation-canyon}) to a conceptual one-dimensional ``bottom of the canyon''. This example illustrates the use of an acquisition function defined {\em on a manifold with lower dimension than the ambient space} in which the original minimization problem was posed.
We propose to approximate this manifold with Geometric Harmonics, and exploit it in accelerating the optimization process via Gaussian processes. One advantage of proceeding in this fashion is that the evaluation/optimization of the lower-dimensional surrogate model can be less expensive than that of the higher-dimensional surrogate one.
We illustrate the use BO when using a surrogate objective function defined by means of Gaussian process regression on a dimensionally-reduced manifold. We only discuss the initial steps, but the scheme converges to the global minimum.

We start by briefly recalling the Bayesian global optimization algorithm~\cite{mockus1989} and we refer the reader to~\cite{shahriari2016,frazier2018} for recent reviews of the topic.
Consider the optimization problem,

\begin{equation}
  \label{eq:optimization-problem}
  \min_{x \in \R^n} f(x),
\end{equation}
where $f$ is the objective function.
The steps of a simple Bayesian optimization scheme for~\eqref{eq:optimization-problem} are presented in Algorithm~\ref{alg:bayesian-optimization}.
\begin{algorithm}
  \caption{Bayesian optimization}
  \label{alg:bayesian-optimization}
  \begin{algorithmic}[H]
    \REQUIRE Gaussian Process prior on $f$. Maximum number of samples $N$.
    \ENSURE Estimated minimizer.
    \STATE{Draw $N_0$ samples from $f$ (from a quasi-random sequence or a Monte Carlo scheme)}.
    \FOR{$\ell = N_0$ \TO $N$}
    \STATE{Update the posterior on $f$ using all available data.}
    \STATE{Minimize the acquisition function $\alpha$ of the current posterior to find the next point $x^{(\ell)}$.}
    \STATE{$y^{(\ell)}$ $\leftarrow$ $f(x^{(\ell)})$}
    \ENDFOR
    \RETURN the point $x$ with smallest $f(x)$ among those evaluated or the point with smallest posterior mean.
  \end{algorithmic}
\end{algorithm}
The algorithm relies on the use of a so-called \emph{acquisition function} $\alpha \colon \R^n \to \R$ that embodies a trade-off between \emph{exploration} and \emph{exploitation} during the algorithm.
Exploration in this context refers to preferentially sampling from regions of the domain that have high posterior variance, whereas exploitation refers to preferentially sampling from regions with lower expected objective value.
Determining new points via the acquisition function and the GP posterior is assumed to be less expensive than evaluating the objective function $f$ directly.

We select the objective function
\begin{equation}
  \label{eq:objective-function}
  f(x_1, x_2) =
  K_1 \left( \arctan\left(\frac{x_2}{x_1}\right) - \frac{\pi}{4} \right)^2
  +
  K_2 \left( x_1^2 + x_2^2 - 1 \right)^2,
\end{equation}
where $K_1$ and $K_2$ are positive constants (specifically, we take $K_1 = 10$ and $K_2 = 50$).
It is straight-forward to see that the global minimum of $f$ is attained at the point $x^\dagger = \frac{\sqrt{2}}{2} (1, 1)$.
We start by drawing $N_0 = 50$ samples in the neighborhood of some initial point (here, $x^{(0)} = (1, 0)$) to obtain the set $X^0 = \{ x^{(\ell)} \in \R^n \, | \, \ell = 1, \dotsc, N_0 \}$.
Gathering $X^0$ can be accomplished through a variety of sampling techniques, e.g. quasi-random sequences~\cite{bratley_programs_1994} or, as in this case, the Metropolis-Hastings algorithm~\cite{metropolis1953} (at a high-enough temperature so that samples are drawn from a sufficiently large vicinity of the initial point).
A few steps of gradient descent on~\eqref{eq:objective-function} suffice to relax the newly-sampled points to the bottom of the canyon (see Figure~\ref{fig:relaxation-canyon}).

Next, we compute the diffusion map embedding $\phi = (\phi_1, \dotsc, \phi_d) \colon \R^n \to \R^d$ of the sampled points and this yields the image $Y^0$ of the sampled points in diffusion map coordinates.
That is,
\begin{equation*}
  Y^0
  =
  \phi(X^0)
  =
  \{ y^{(\ell)} = \phi(x^{(\ell)}) \in \R^d \, | \, \ell = 1, \dotsc, N_0 \}.
\end{equation*}
Note that the canonical distribution associated to our objective function $f$ is concentrated around $M$, the positive orthant of the two-dimensional sphere of radius one,
\begin{equation*}
  M
  =
  \left\{
    x = (x_1, x_2) \in \R_{\ge 0} \times \R_{\ge 0} \, | \, x^\top x = 1
  \right\}.
\end{equation*}
Consequently, we expect that a single diffusion map coordinate will suffice for embedding the slow manifold of the optimization algorithm dynamics; in other words, we assume that $d = 1$ in $\phi = (\phi_1, \dotsc, \phi_d)$.
Indeed, it can be shown that a diffusion map coordinate on points sampled from the probability measure proportional to $\mathrm{e}^{-\beta f(x)} \mathrm{d} x$ is one-to-one with the arc-length of $M$.

Next, we apply Gaussian Process Regression to the labeled dataset
\begin{equation*}
  \{ (y^{(\ell)}, f(x^{(\ell)})) \in \R^d \times \R \, | \, \ell = 1, \dotsc, N_0 \}
\end{equation*}
to obtain its mean $g(y)$, which we will regard as a surrogate objective function, and its standard deviation $\sigma(y)$.
We then construct the acquisition function
\begin{equation*}
  \alpha(y)
  =
  g(y)
  -
  \kappa \, \sigma(y)
  +
  \tau \, \frac{\dist(y, Y^0)}{m_0},
\end{equation*}
where $\kappa, \tau > 0$ are constants (taken in our case to be, respectively, equal to 1.96 and 3), $m_0$ denotes the median Euclidean distance between pairs of points in $Y^0$, and $\dist(y, Y^0)$ is the distance between $y$ and the set $Y^0$ or, equivalently,
\begin{equation*}
  \dist(y, Y^0) = \min_{y^{(\ell)} \in Y^0} \| y - y^{(\ell)} \|.
\end{equation*}
Observe that minimizing the acquisition function $\alpha$ amounts in practice to finding a new point $y^\star$ that strikes a balance between minimizing the surrogate objective function $g$, belonging to a region of high-uncertainty (via the $\kappa$ penalty), or lying far away from the sampled points in the embedding space (via the $\tau$ penalty).

Thus, once we find $y^\star = \argmin \alpha$, we can choose an arbitrary point $x^{(1)} \in \R^n$ within the preimage
\begin{equation*}
  \phi^{-1}(\{ y^\star \})
  =
  \left\{
    x \in \R^2 \, | \, \phi(x) = y^\star
  \right\}.
\end{equation*}
The preimage can be thought of as blob or point-cloud that maps to the specified point $y^\star$ in diffusion map space.
The construction of points in the set $\phi^{-1}(\{ y^\star \})$ is called lifting in the equation-free literature~\cite{vandekerckhove2011}.
One practical way of lifting points consists of sampling points from the measure determined by the potential function
\begin{equation*}
  \eta f(x) + \nu \| \phi(x) - y^\star \|^2,
\end{equation*}
for some constants $\eta \ge 0$ and $\nu > 0$.
In this particular case, however, we lift the point $y^\star$ to $x^{(1)}$ by constructing, via GP regression, a function that maps $y^{(\ell)}$ to $x^{(\ell)}$ for $\ell = 1, \dotsc, N_0$.
Doing so, allows us to obtain not just a point estimate for $x^{(1)}$ but also an estimate for the uncertainty of its coordinates (see the orange regions at the base of Figure~\ref{fig:gp-curved-domain-illustration}).

Next, we proceed analogously as we did with $x^{(0)}$.
We can sample $N_1$ points in the neighborhood of $x^{(1)}$ to obtain
\begin{equation*}
  X^1
  =
  X^0
  \cup
  \{ x^{(\ell^\prime)} \in \R^n \, | \, \ell^\prime = 1, \dotsc, N_1 \}.
\end{equation*}
This new dataset in turn yields another set of diffusion map coordinates $Y^1$, another GP regression, \textrm{etc.}
In the simplest case, we just take $N_i = 1$ for $i > 0$ and obtain $X^{i}$ by adjoining the point $x^{(i-1)}$ to $X^{i-1}$.

We now have an iterative process to go back and forth between the original domain of the objective function $f$ and the dimensionally-reduced domain of the surrogate function $g$. The obvious problem-dependent issues (how frequently do we streamingly update $\phi$, how we test for changes in dimension of the reduced manifold, etc.) are outlined in section~\ref{sec:dimension reduction} and are clearly the subject of future work.
An interesting avenue of research would be to study how to optimally recompute the dimensionality reduction step so that it requires the least amount of computational effort.
Possible sources of ideas could come from landmarking as in~\cite{gao2019a} or~\cite{long2017}, as well as from streaming PCA algorithms~\cite{huang2021}, or from~\cite{giannakis2021}.
The potential advantage of calculating diffusion map coordinates and working in the reduced space is that (a) the evaluation of the surrogate is cheaper in the reduced space, but also (b) the number of coordinates in which this surrogate is optimized is lower.
Standard BO consists of the same iterative procedure that we have described, but replacing the embedding $\phi$ by the identity mapping (\emph{i.e.,} we set $d = n$ and the embedding $\phi = \id \colon \R^n \to \R^n$ simply maps every point to itself).


\end{document}